\documentclass{article}

\usepackage{arxiv}

\usepackage[utf8]{inputenc} 
\usepackage[T1]{fontenc}    
\usepackage{hyperref}       
\usepackage{url}            
\usepackage{booktabs}       
\usepackage{amsfonts}       
\usepackage{nicefrac}       
\usepackage{microtype}      
\usepackage{amsmath}
\usepackage{cleveref}       
\usepackage{lipsum}         
\usepackage{graphicx}
\usepackage{natbib}
\usepackage{doi}
\usepackage{tabularx}
\usepackage{multirow}

\usepackage{floatrow} 
\title{MFAF: An EVA02-Based Multi-scale Frequency Attention Fusion Method for Cross-View Geo-Localization}

\date{}

\newif\ifuniqueAffiliation
\uniqueAffiliationtrue

\ifuniqueAffiliation 
\author{ 
  LIU YiTong \qquad    
  LIU TianZhu \qquad  
  GU YanFeng$^*$      
  \ \\ 
  \ School of Electronics and Information Engineering, Harbin Institute of Technology, Harbin 150001, China
}
\else
\usepackage{authblk}

\newbox{\orcid}\sbox{\orcid}{\includegraphics[scale=0.06]{orcid.pdf}} 
\author[1]{LIU YiTong\thanks{\texttt{yitong.liu@hit.edu.cn}}}%
\author[1]{LIU TianZhu\thanks{\texttt{tianzhu.liu@hit.edu.cn}}}%
\author[1]{GU YanFeng$^*$\thanks{\texttt{yanfeng.gu@hit.edu.cn ($^*$Corresponding author)}}}%
\affil[1]{School of Electronics and Information Engineering, Harbin Institute of Technology, Harbin 150001, China}
\fi

\renewcommand{\shorttitle}{MFAF for Cross-View Geo-Localization}

\hypersetup{
pdftitle={MFAF: An EVA02-Based Multi-scale Frequency Attention Fusion Method for Cross-View Geo-Localization},
pdfsubject={cs.CV (Computer Vision and Pattern Recognition)}, 
pdfauthor={LIU YiTong, LIU TianZhu, GU YanFeng},
pdfkeywords={Cross-View Geo-Localization, EVA02, Attention Fusion, Multi-scale Frequency},
}

\begin{document}
\maketitle

\begin{abstract}
	Cross-view geo-localization aims to determine the geographical location of a query image by matching it against a gallery of images. This task is challenging due to the significant appearance variations of objects observed from variable views, along with the difficulty in extracting discriminative features. Existing approaches often rely on extracting features through feature map segmentation while neglecting spatial and semantic information. To address these issues, we propose the EVA02-based Multi-scale Frequency Attention Fusion (MFAF) method. The MFAF method consists of Multi-Frequency Branch-wise Block (MFB) and the Frequency-aware Spatial Attention (FSA) module. The MFB block effectively captures both low-frequency structural features and high-frequency edge details across multiple scales, improving the consistency and robustness of feature representations across various viewpoints. Meanwhile, the FSA module adaptively focuses on the key regions of frequency features, significantly mitigating the interference caused by background noise and viewpoint variability. Extensive experiments on widely recognized benchmarks, including University-1652, SUES-200, and Dense-UAV, demonstrate that the MFAF method achieves competitive performance in both drone localization and drone navigation tasks.
\end{abstract}


\section{Introduction}
With the rapid development of drone technology and the continuous reduction in production costs, drones have found widespread application, such as aerial photography\cite{aerial_photography}, geological exploration\cite{geological_xploration} and disaster response\cite{disaster_response}. In these cases, drones leverage their precise real-time data collection to support environmental monitoring and resource management. However, drone operation is dependent on the support of positioning systems. Geo-localization typically relies on the Global Positioning System (GPS), but GPS signals are susceptible to interference from factors such as weather and buildings. In some complex environments, GPS signals may be completely lost, preventing them from providing accurate location information. Cross-View Geo-Localization (CVGL) in Unmanned Aerial Vehicle (UAV) scenarios, as an emerging positioning method, can precisely determine the geographic location of a query image by matching it with reference images marked with GPS coordinates, without relying on GPS signals. 

CVGL typically relies on visual algorithms to mitigate environmental interference, with the objective of extracting unified high-dimensional feature representations of image pairs and evaluating their similarity. However, due to variations in the perspective angles between drone and satellite imagery, images of the same location often appear significantly different. Therefore, how to effectively reduce the feature differences between images taken from variable views has become one of the main challenges. Additionally, for large-scale CVGL systems, the design of adapters to extract highly discriminative image features is subject to higher requirements. The adapter must comprehensively consider both geometric and semantic information between image pairs to enhance the discriminative capability of feature representations. Given the flexibility and diversity of drone flight trajectories, images captured by drones often experience variations in shooting angles and heights. As a result, the CVGL method must be adaptable to these changes, ensuring the extraction of globally robust features.

\begin{figure}[h]
\centering
\includegraphics[width=\textwidth, keepaspectratio=true]{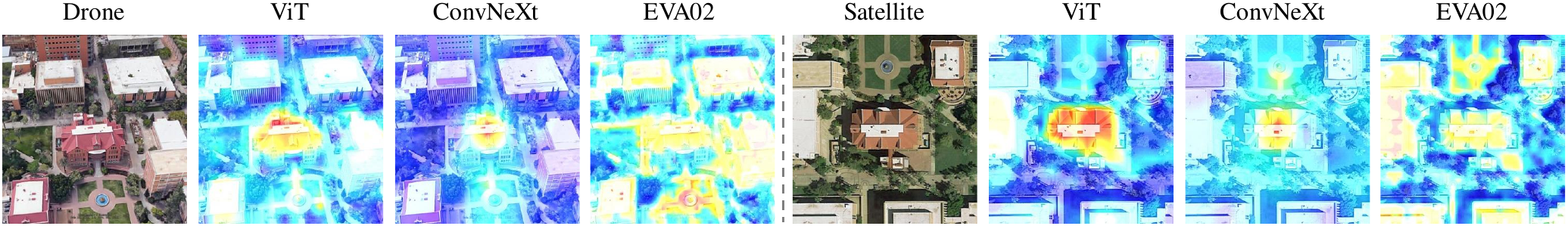}
\caption{The heatmaps generated by backbones used in CVGL methods. The ``ViT'' column illustrates heatmaps generated by the ViT backbone used in FSRA method. ``ConvNeXt'' column illustrates heatmaps generated by the ConvNeXt backbone used in MCCG method. ``EVA02'' column illustrates heatmaps generated by the EVA02 backbone.}
\label{fig：热力图}
\end{figure}

Current feature extraction methods for CVGL predominantly rely on CNNs \cite{LPN,GeoDTR,Joint,MCCG} and Vision Transformers (ViTs) \cite{FSRA,SDPL,SRLN}. CNNs capture local features through receptive fields and hierarchical pooling. However, their locality and down-sampling often result in the loss of fine spatial details and an overemphasis on central regions, limiting global spatial modeling. While traditional ViTs offer global receptive fields but lack explicit spatial positional encoding, which results in attention bias toward central areas and insufficient modeling of peripheral structural information. EVA02 \cite{EVA02}, as an enhanced visual foundation model, addresses these limitations by incorporating explicit spatial relationship modeling, thereby improving global layout perception. As illustrated in Figure \ref{fig：热力图}, attention maps from FSRA’s ViT and MCCG’s ConvNeXt exhibit central focus, failing to contextualize surrounding regions. In contrast, EVA02 leverages global information and an optimized attention mechanism to effectively distinguish buildings from surrounding vegetation and background, significantly improving image matching performance.

Meanwhile, existing CVGL methods typically learn feature representations by segmenting the entire image content or feature maps, but these methods fail to fully capture deep semantic information, leading to information loss. For example, methods such as LPN\cite{LPN} and FSRA\cite{FSRA} perform segmentation on the last feature map of the input image, which may disrupt the structural integrity of objects in the image, resulting in the loss of semantic information. Another multi-classification method, MCCG\cite{MCCG}, extracts features through cross-dimensional interactions between the last feature map and the spatial and channel dimensions of the input, generating multiple feature representations to enable image matching and scene classification across different views. 

To address the aforementioned issues, we propose a new method called MFAF, which stands for the EVA02-based Multi-scale Frequency Attention Fusion Method for CVGL. To address the limitation of existing methods, where backbones tend to focus on the central regions of images while neglecting the edges and overall spatial structure, the proposed MFAF method incorporates the visual foundation model EVA02. This enhances the capacity to capture global semantic information, enabling the extraction of comprehensive global features from images of different views. To further capture fine-grained features, MFAF integrates two modules: the Multi-Frequency Branch-wise Block (MFB) and the Frequency-aware Spatial Attention (FSA) module. The MFB block utilizes a multi-frequency branch structure, where the Multi-scale Low-Frequency Branch captures relatively stable structural information across different views, facilitating the establishment of structural correspondences between images from varying perspectives. In contrast, the Hybrid High-Frequency Branch retains intricate texture details, improving the sensitivity to local discrepancies. In addition, the FSA module adaptively focuses on key regions within different frequency features through a frequency-aware mechanism, augmenting the representation of salient features and optimizing the discriminative power and generalizability of the overall feature set. Moreover, to enable more accurate geographic label predictions, this study introduces the Multi-Classifier Block (MCB)\cite{MCCG}, in conjunction with the Cross-Entropy Loss and Cross-Domain Triplet Loss\cite{FSRA}, significantly enhancing the accuracy and robustness of CVGL.

In summary, the main contributions of this paper include the following:

(1) A novel MFAF module is proposed that effectively exploits frequency information, which inherently encapsulates both structural patterns and fine-grained texture details by designing the MFB and FSA modules. This optimization enhances the feature representation and discrimination capabilities of cross-view images by capturing both spatial and spectral characteristics, thereby improving robustness to complex viewpoint variations.

(2) The visual foundation model EVA02 is innovatively introduced to the CVGL task and integrated with the MFAF module. This fusion improves the dual ability of the method to model global semantic information and local detail features, improving its adaptability and performance on large-scale datasets.

(3) On three public datasets, University-1652, SUES-200 and Dense-UAV, the proposed MFAF method achieved optimal performance in drone localization and drone navigation tasks. The effectiveness and robustness of the method were validated through comparative experiments and ablation studies.

The rest of the paper is organized as follows. In Section 2, we discuss related work. Section 3 describes the proposed MFAF method in detail. Section 4 presents the experimental results and analysis of the ablation study on three benchmarks. Finally, the conclusion is drawn in Section 5.

\section{Related work}
CVGL as a subset of image retrieval focuses on extracting high discriminative feature representations. Traditional methods based on pixel matching\cite{pixel_matching} and manual feature extraction\cite{manual_feature} struggle to adapt to significant view changes and view occlusions, leading to issues such as low computational efficiency and poor matching accuracy. In recent years, deep neural network feature extraction methods have gained prominence in cross-view tasks by automatically learning discriminative high-level semantic features.

\textbf{(1) Feature Extraction}. Network feature learning methods use deep architectures and metric learning to effectively handle images from variable views. Workman et al.\cite{Workman} first introduced CNNs to learn joint semantic feature representations for cross-view tasks and proposed the large-scale dataset CVUSA consisting of ground and satellite imagery. Furthermore, Deuser et al.\cite{Deuser} used the ConvNeXt network to extract image features and used geographical proximity and visual similarity to identify challenging samples, significantly improving localization performance. Although CNNs have made significant progress in modeling appearance, these methods struggle to comprehensively capture the global information required for cross-view tasks. In recent years, ViTs\cite{Transformer,VisionTrans} have gained attention due to the self-attention mechanism, which enhances global contextual reasoning capabilities and overcomes the limitations of CNNs. Yang et al.\cite{Yang} were the first to introduce ViTs into cross-view tasks and proposed a self-cross attention mechanism, significantly enhancing feature differences between different network layers. Additionally, Zhao et al.\cite{Zhao} proposed the Mutual Generative Transformer and introduced the Cascaded Attention Masking strategy, further improving the localization accuracy. With the rise of large-scale visual foundation models, cross-view tasks have increasingly adopted these models as the initial backbone during training. Self-supervised visual encoders such as DINO\cite{DINO}, DINOv2\cite{DINOv2}, EVA\cite{EVA}, and EVA02 have become essential tools in this field. Although these foundation models demonstrate good performance on various visual tasks, they still face performance degradation challenges in cross-view tasks due to significant distribution differences from variable views.

\textbf{(2) Training Strategy Optimization}. Model training is central to deep learning, and much research is focused on improving and designing effective loss functions. In cross-view tasks, deep metric learning employs triplet loss\cite{triplet_loss}, instance loss\cite{instance}, contrastive loss\cite{Workman}, and their various extensions, such as distance-based logistic loss\cite{distance_based}, weighted soft margin ranking loss\cite{weighted_soft_margin}, and hard exemplar reweighting triplet loss\cite{hard_exemplar_reweighting}.In the application of triplet loss, it is crucial to carefully sample triplets. In the absence of advanced techniques, the inclusion of hard negative samples may lead to model collapse. 

\section{Method}
\subsection{Overall architecture}
\begin{figure}[!t]
\centering
\includegraphics[width=\textwidth, keepaspectratio=true]{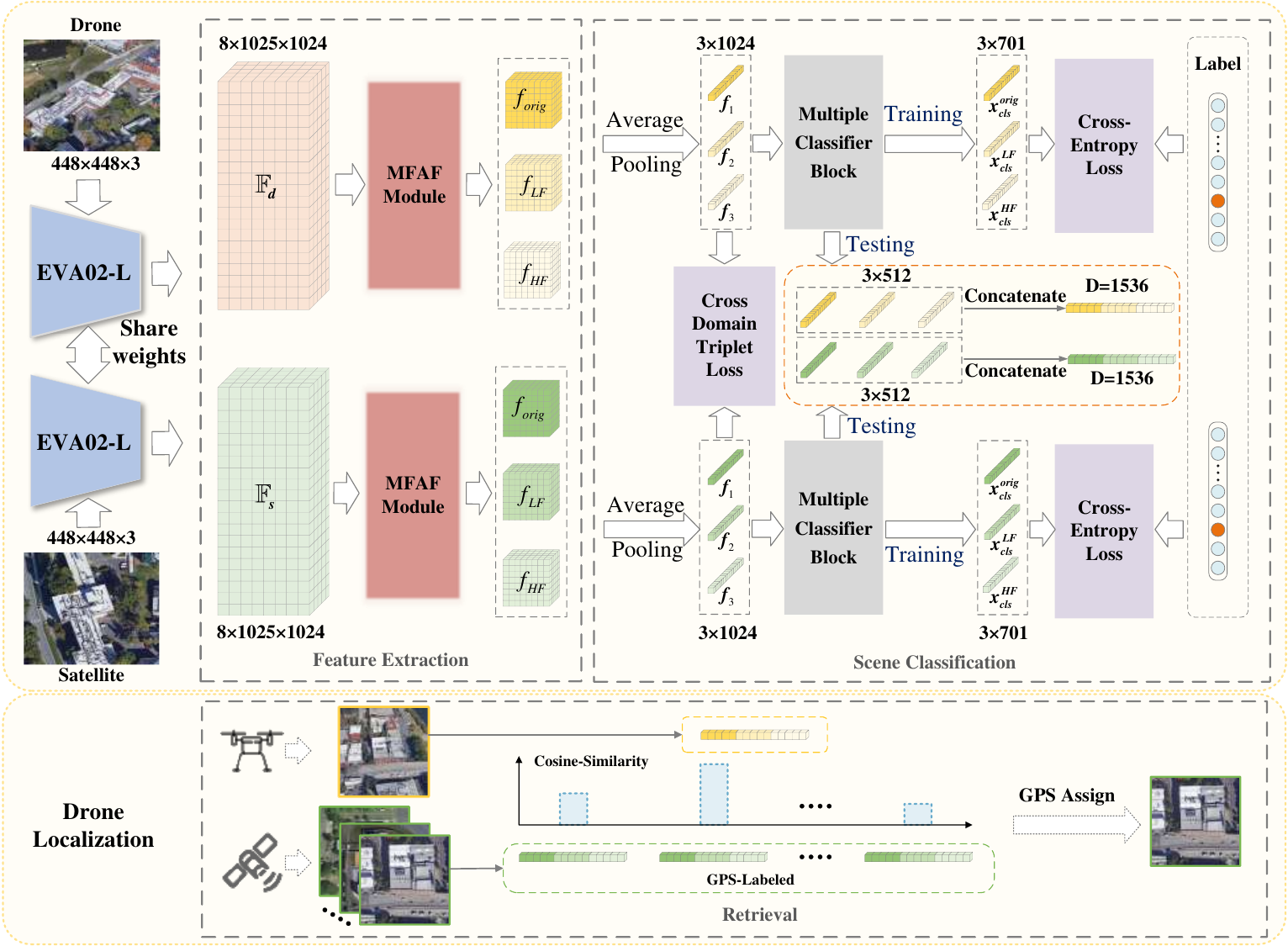}
\caption{The pipeline of proposed MFAF method and drone localization task. Feature extraction comprises a pair of EVA02-L backbone with shared weights and our MFAF module. Scene classification introduces MCB module and utilizes both cross-domain triplet loss and cross-entropy loss.}
\label{fig：总体}
\end{figure}

Similarly to the Siamese network, our method adopts a dual-branch architecture corresponding to drone and satellite views, and branches share network weights. Each branch comprises two key components: feature extraction and supervised classification. As illustrated in Figure \ref{fig：总体}, the feature extraction component integrates the EVA02 backbone for fundamental feature encoding and the MFAF module. The MFAF module is designed to enhance the information across different frequency bands within the extracted features. The scene classification component is composed of an MCB module coupled with cross-entropy loss function and cross-domain triplet loss. Finally, the classification features extracted by the MCB module are used to compute the cosine similarity between the query (e.g., drone-view) and gallery (e.g., satellite-view) features. The satellite GPS corresponding to the highest similarity is selected as the drone's GPS, thereby achiving cross-view geo-localization.

\subsection{Feature extraction}
\subsubsection{Basic feature extraction}
In the CVGL task, given a dataset $D=\left\{\left(x_v^i, y_i\right) \mid i=1,2, \cdots, N\right\}$, where $x_v^i$ represents the input images, $v \in[d, s]$ indicates the different views, $x_d^i$ corresponds to the drone view images and $x_s^i$ corresponds to the satellite view images. The corresponding label of the category of the scene is represented by $y_i$, and $N$ denotes the total number of categories. EVA02 model as the feature extraction module, aiming to transform drone and satellite images into high-dimensional feature representations. Specifically, given an input image $x_v^i$ of size $\mathit{H} \times \mathit{W} \times \mathit{C}$, the feature mapping function $E(\cdot)$ produces a feature map $\mathbb{F}_v \in \mathbb{R}^{H \times W \times C}$, which is defined as follows:
\begin{equation}
\mathbb{F}_v=E\left(x_v^i\right)
\label{eq1}
\end{equation}
In this context, $\mathbb{F}_d=E\left(x_d^i\right)$, $\mathbb{F}_s=E\left(x_s^i\right)$, $\mathbb{F}_d$ and $\mathbb{F}_s$ represent the features extracted from the drone and satellite view images. For instance, when the input consists of image pairs of size $448 \times 448 \times 3 \quad(\mathit{H} \times \mathit{W} \times \mathit{C})$, $E\left(x_v^i\right)$ generates two feature maps of size $8 \times 1025 \times 768$ each.

\begin{figure}[!t]
\centering
\includegraphics[width=0.75\textwidth, keepaspectratio=true]{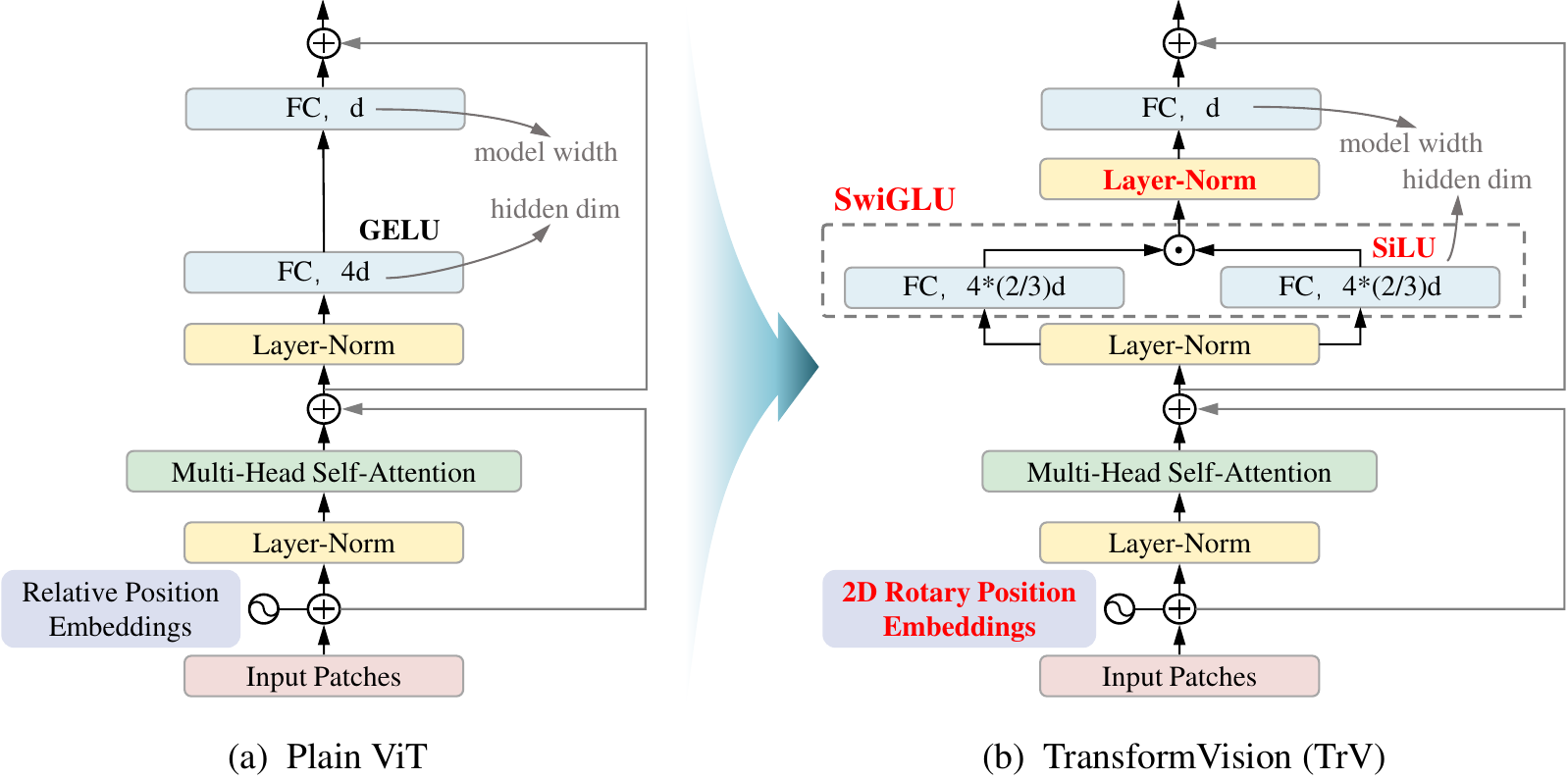}
\caption{Explanation of the blocks in VisionTransformer (ViT) and the TransformVision (TrV) proposed in EVA02. (a) Plain ViT block; (b) TrV block: TrV builds on the plain ViT and incorporates functions like SwiGLU, sub-LN, and 2D RoPE. To maintain parameter consistency, the hidden dimension of SwiGLU is set to 2/3 of the typical MLP size.}
\label{fig:EVA02}
\end{figure}

EVA02 is a Transformer-based visual foundation model that utilizes a Plain ViT architecture with optimizations, introducing the concept of Transform Vision (TrV). The basic architecture is depicted in Figure \ref{fig:EVA02} (b). In the TrV blocks, the traditional MLP$+$GELU\cite{GELU} in ViTs is replaced by SwiGLU (Sigmoid-Gated Linear Unit)\cite{SwiGLU}. To maintain consistency in both parameters and FLOPs, TrV sets the hidden dimension of the position-wise feed-forward network (FFN) to two-thirds of the conventional MLP. Additionally, TrV employs the Xavier Normal initialization strategy\cite{Xavier} to mitigate potential performance degradation caused by the random initialization of the SwiGLU. TrV adopts a sub-LN architecture (MHSA→LN→FFN), as opposed to the pre-LN structure used in ViTs\cite{LN,sub_LN}. This adjustment alleviates the vanishing gradient problem in deep networks and strengthens the stability of gradient flow. Contrasting with absolute or standard relative positional embeddings in ViTs, TrV implements a 2D RoPE (Relative Positional Encoding) scheme\cite{RoPE,SwinTransformerV2}, preserving the ability to model spatial relationships while avoiding instability during pretraining.

\subsubsection{MFAF module}
\begin{figure}[!t]
\centering
\includegraphics[width=\textwidth, keepaspectratio=true]{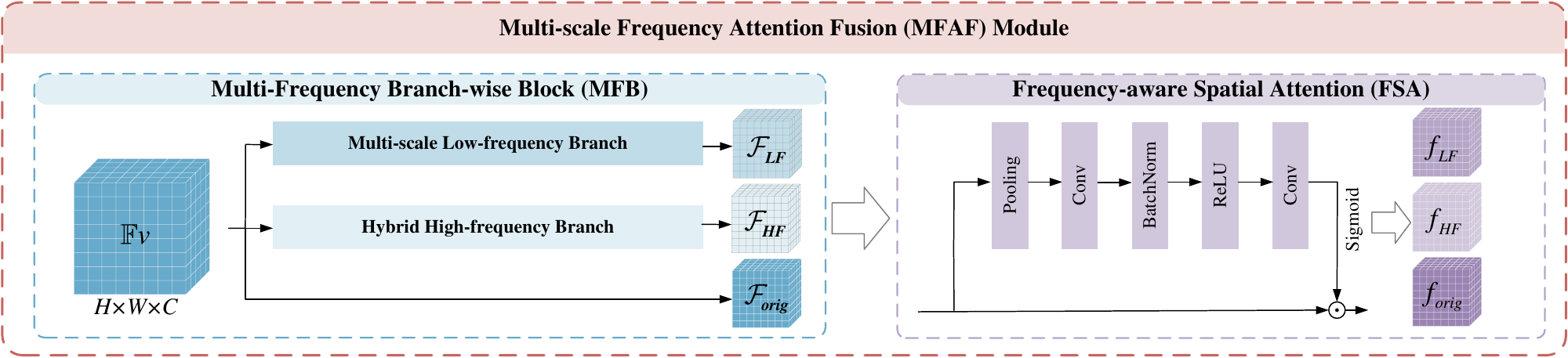}
\caption{Illustration of the MFAF module,which consists of a Multi-Frequency Branch-wise Block (MFB) and a Frequency-aware Spatial Attention (FSA) mechanism.}
\label{fig:MFAFmodule}
\end{figure}

\begin{figure}[!t]
\centering
\includegraphics[width=\textwidth, keepaspectratio=true]{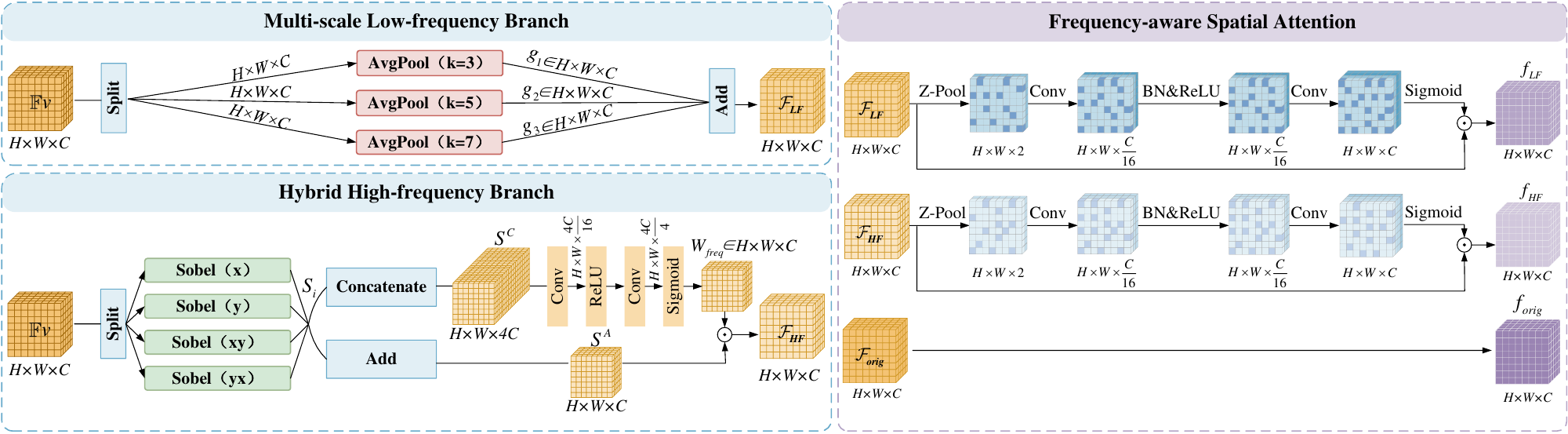}
\caption{Architectural of the MFB and FSA modules.The left side shows the principles of the Multi-scale Low-frequency Branch and Hybrid High-frequency Branch in the MFB block. And the right side illustrates the attention mechanism of the FSA module.}
\label{fig:MFAF}
\end{figure}

The significance of frequency information in feature extraction has been widely validated in the field of image processing. Abello et al.\cite{Abello} discovered through spectral block experiments that networks exhibit distinct dependencies on frequency information. Specifically, shallow networks primarily rely on mid-to-high frequency information, whereas deeper networks progressively emphasize low-frequency global semantic features. The F3-Net model proposed by Sun et al. \cite{F3_Net}, designed for UAV-based geo-localization, further supports this finding. Their study demonstrates that shallow networks primarily capture texture features, which are closely linked to high-frequency components, while deeper layers are more adept at extracting low-frequency features, ultimately facilitating the development of a model that can accurately capture fine-grained image details. Inspired by these studies, we propose a novel module called MFAF, which integrates the Multi-Frequency Branch-wise Block with the Frequency-aware Spatial Attention module (as illustrated in Figure \ref{fig:MFAFmodule} and \ref{fig:MFAF}). The MFAF module aims to extract deep frequency features while leveraging key semantic information within the existing network structure, thereby enhancing the multi-view invariance of the features. Through consistent representation of salient target semantic features, MFAF ultimately improves the robustness of multi-view image features.

The principle of the proposed method is as follows:
Given an input feature map $\mathbb{F}_v \in \mathbb{R}^{H \times W \times C}$, after passing through the Multi-scale Low-frequency Branch and the Hybrid High-frequency Branch, the features $\mathcal{F}_{L F} \in \mathbb{R}^{H \times W \times C}$ emphasizing low-frequency information and high-frequency information $\mathcal{F}_{H F} \in \mathbb{R}^{H \times W \times C}$ are obtained respectively. The Multi-scale Low-frequency Branch extracts hierarchical structural information at different scales through average pooling. The mathematical expression for this process is:
\begin{equation}
g_i=\operatorname{AvgPool}_{k_i}\left(\mathbb{F}_v\right) \otimes \boldsymbol{\omega}_i, \quad k_1=3, k_2=5, k_3=7
\end{equation}
where $\operatorname{AvgPool}_{k_i}$ denotes the average pooling operation with a kernel size of $k_i$ , and $\omega_i$ represents the learnable channel weight parameters, while $\otimes$ indicates the multiplication operation per channel. The output of the multi-scale pooling is fused through a linear combination, with the specific formula:
\begin{equation}
\mathcal{F}_{L F}=\sum_{\mathrm{i}=1}^3 g_{\mathrm{i}}
\end{equation}
The design forms a hierarchical representation of structural information from local to global: $3 \times 3$ pooling captures small-scale texture patterns, $5 \times 5$ pooling integrates mid-scale structural information, and $7 \times 7$ pooling aids in capturing global semantic layouts, thereby enabling multi-scale structural representation.

In the Hybrid High-frequency Branch, a four-direction Sobel operator is employed to extract edge-related high-frequency features. The frequency response characteristics of each directional operator are as follows:
\begin{equation}
K_x=\left[\begin{array}{ccc}
-1 & 0 & 1 \\
-2 & 0 & 2 \\
-1 & 0 & 1
\end{array}\right], \quad K_y=\left[\begin{array}{ccc}
-1 & -2 & -1 \\
0 & 0 & 0 \\
1 & 2 & 1
\end{array}\right], \quad K_{x y}=\left[\begin{array}{ccc}
-2 & -1 & 0 \\
-1 & 0 & 1 \\
0 & 1 & 2
\end{array}\right], \quad K_{y x}=\left[\begin{array}{ccc}
0 & 1 & 2 \\
-1 & 0 & 1 \\
-2 & -1 & 0
\end{array}\right]
\end{equation}
Applying these operators to the input feature map $\mathbb{F}_v$ through grouped convolutions, we obtain edge responses with directional selectivity:
\begin{equation}
\mathrm{S}_i=\left|\operatorname{Conv}\left(\mathbb{F}_v, K_i\right)\right|, \quad i \in\{x, y, x y, y x\}
\end{equation}
To enhance the discrimination of edge features, the edge responses of the four directions are concatenated into a single channel feature $S^C \in \mathbb{R}^{H \times W \times 4C}$. Subsequently, a convolution operation is used to reduce and expand the dimensions, generating channel weights $W_{\text {edge }}$, which are element-wise multiplied with the directional edge feature maps, resulting in the high-frequency attention features $\mathcal{F}_{H F} \in \mathbb{R}^{H \times W \times C}$:
\begin{equation}
\mathcal{F}_{H F}=W_{\text {edge }} \odot S^A
\end{equation}

To further enhance the discriminative capacity of frequency features, we propose the Frequency-aware Spatial Attention module. The processing is as follows: both the mean and maximum pooling of the frequency feature map are computed simultaneously, generating two statistical channels:
\begin{equation}
z=[\operatorname{AvgPool}(\mathcal{F}), \operatorname{MaxPool}(\mathcal{F})]
\end{equation}
Next, these two channel features are reduced in dimension using convolution and then expanded to match the number of channels in the input feature map. The final attention weights are generated using the Sigmoid activation function:
\begin{equation}
W_{\text {freq }}=\sigma\left(W_2 \bullet \operatorname{ReLU}\left(W_1 \bullet z\right)\right)
\end{equation}
where $W_1 \in \mathbb{R}^{(C / \text { reduction }) \times 2 \times 1 \times 1}$, and $W_2 \in \mathbb{R}^{C \times(C / \text { reduction }) \times 1 \times 1}$ represents the learnable spatial weight parameters. Ultimately, the features containing frequency information are obtained by:
\begin{equation}
f_m=S^A \odot W_{\text {freq }}
\end{equation}
where different feature branches are represented by $m \in\{LF, HF\}$.

Additionally, in the FSA input stage, the original input feature map is preserved and denoted as $\mathcal{F}_{\text {orig }} \in \mathbb{R}^{H \times W \times C}$, while in the FSA output stage, it is represented as  $f_{\text {orig }} \in \mathbb{R}^{H \times W \times C}$.

Overall, the output of the MFAF module consists of features from three branches: 
\begin{equation}
f_m \in \mathbb{R}^{H \times W \times C},\quad m \in\{orig, LF, HF\}
\label{eq:m}
\end{equation}

\subsection{MCB module}
\begin{figure}[!t]
\centering
\includegraphics[width=0.7\textwidth, keepaspectratio=true]{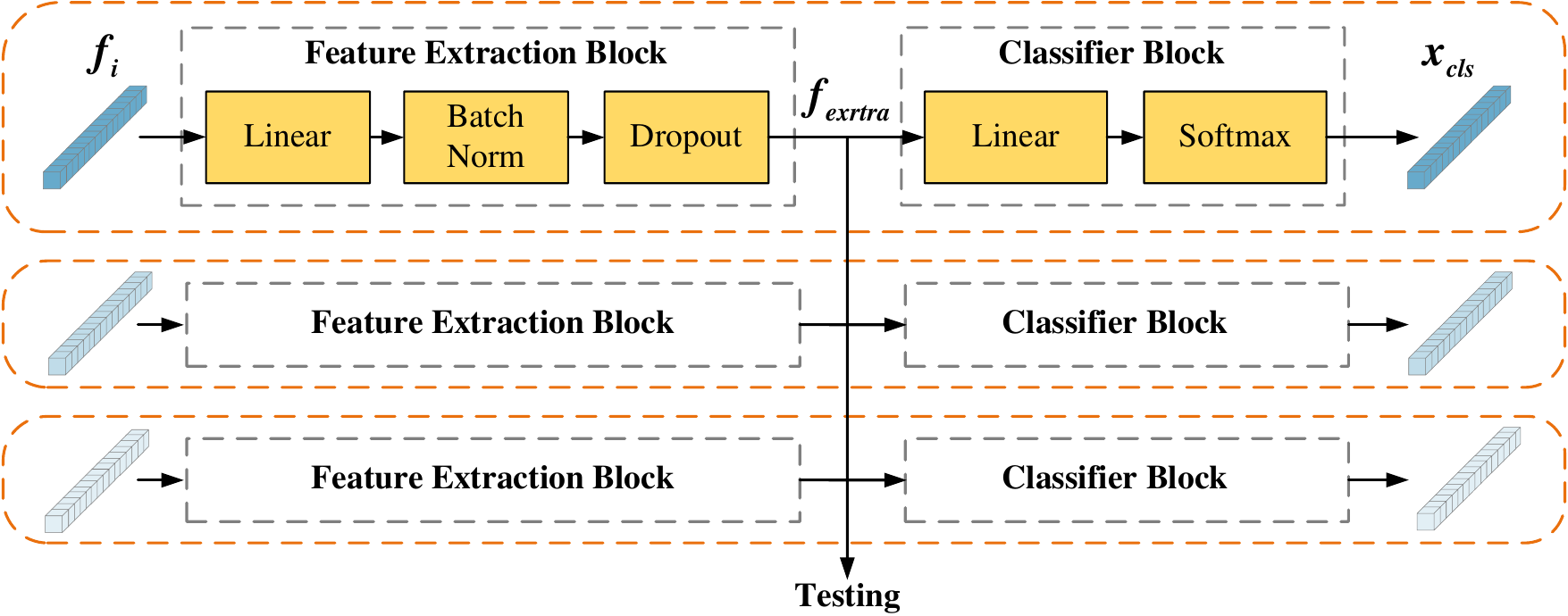}
\caption{Process of the MCB module.The feature vectors after average pooling, pass through three Feature Extraction Blocks to generate three feature vectors for testing. Then, the feature vectors are processed by Classifier Block to produce scene categories.}
\label{fig:MCB}
\end{figure}

The proposed method introduces the Multi-Classifier Block (MCB) during the network training process to fully exploit the information within the image and increase distinguish between different instances. As illustrated in Figure \ref{fig:MCB}, the input to each branch of the MCB is the feature vector $f_m$ obtained through the MFAF module, which undergoes average pooling to generate the feature vector $f_j(j=1,2,3)$. These feature vectors are then passed through a multi-classifier module consisting of three classifier layers, yielding distinctive feature representations. During the training, each feature vector sequentially passes through the feature extraction and classifier modules, being transformed into class vectors $x_{c l s}^{L F}$, $x_{c l s}^{H F}$ and $x_{c l s}^{\text {orig }}$, which optimize based on cross-entropy loss function. During testing, the model only utilizes the feature extraction module, generating feature representations $f_{\text {exta }}^{L F}$, $f_{\text {exta }}^{H F}$ and $f_{\text {exta }}^{orig}$ of dimension $3 \times 512$. These three features are then concatenated into a 1536-dimensional feature vector, which is used for drone localization and drone navigation tasks.

\subsection{Loss function}
In CVGL task, each scene can be regarded as an independent category, and accurately classifying and matching features across different views is the challenge. We employ the cross-entropy loss function in this paper. Its mathematical expression is as follows:
\begin{equation}
L_{C E}=-\sum_{i=1}^N p\left(y_i\right) \log q\left(x_{c l s}^i\right)
\end{equation}
where $p\left(y_i\right)$ represents the true probability that the sample belongs to class $y_i$, and $q\left(x_{c l s}^i\right)$ denotes the predicted probability of the model that the sample belongs to class $y_i$. Since the true label is represented as a one-hot vector, let the true label of the sample be $y_{\text {true }}$, and the predicted class probability distribution be $\hat{x}$. The cross-entropy loss can be simplified to:
\begin{equation}
L_{C E}=-\sum_{i=1}^N y_{\text {true }}^i \log \left(\hat{x}^i\right)
\end{equation}
\begin{figure}[h]
\centering
\includegraphics[width=\textwidth, keepaspectratio=true]{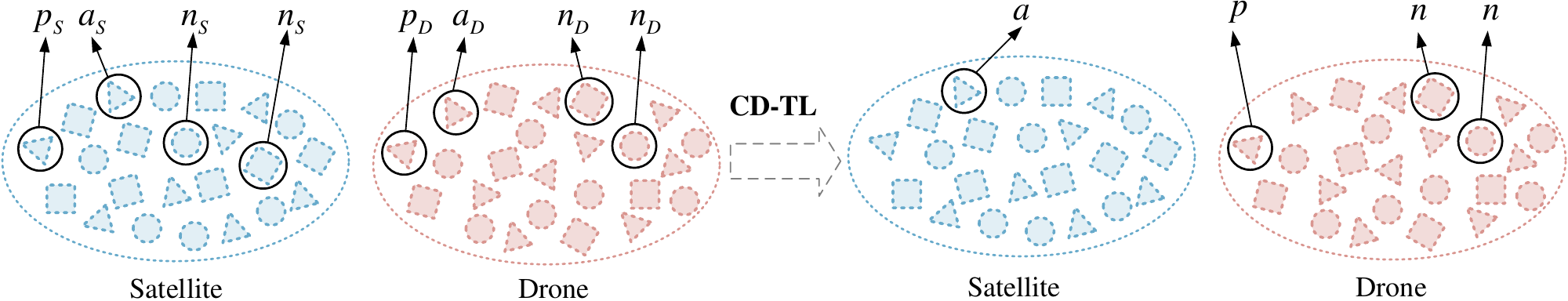}
\caption{The composition of the image pairs changes after using the cross-domain triplet loss. The same color indicates images from the same domain, and different shapes represent images from different scene categories.}
\label{fig:CD-TL}
\end{figure}
For CVGL, significant discrepancies in data distribution across different domains pose a formidable challenge to the model generalization capability. To address this issue, the Cross-Domain Triplet Loss (CD-TL) is introduced as an effective loss function, with the objective of narrowing the distributional gap between domains and guiding the model to learn feature representations with cross-domain consistency. In traditional triplet loss, each triplet consists of anchor, positive and negative:
\begin{equation}
L_{\text {triplet }}=\max (0, d(a, p)-d(a, n)+M)
\end{equation}
where $a$ denotes the anchor, $p$ the positive and $n$ the negative. $d(\cdot, \cdot)$ represents the distance function, and $M$ refers to the aforementioned hyperparameter, which is set to 0.3 in our experiments. In the context of cross-domain learning, the CD-TL further extends this concept. As illustrated in Figure \ref{fig:CD-TL}, our task involves two distinct viewpoint domains, drone-view domain $D$ and satellite-view domain $S$, with their respective data distributions denoted as $P_D(x)$ and $P_S(x)$. The model aims to learn a feature mapping $f(\cdot)$ that ensures similar samples from both the drone and satellite image domains remain close, while dissimilar samples are pushed further apart. For example, we select a sample from the drone-view domain as the anchor $a_D$, a sample from the satellite-view domain with the same class label as the positive $p_S$, and a sample from a different class in the satellite domain as the negative $n_S$. The cross-domain triplet loss function is then expressed as:
\begin{equation}
L_{\mathrm{CD} \text {-triplet }}=\max \left\{0, d\left[f\left(a_D\right), f\left(p_S\right)\right]-d\left[f\left(a_D\right), f\left(n_S\right)\right]+M\right\}
\end{equation}

During the training, the total loss consists of two components: the cross-entropy loss and the cross-domain triplet loss. The overall loss is computed as follows:
\begin{equation}
L=\frac{1}{3} \sum_{j=1}^3\left(L_{\mathrm{CE}}^j+L_{\mathrm{CD} \text {-triplet }}^j\right)
\end{equation}
where ``3'' indicates the total number of branches in the MFSA module, and $j$ refers to the $m$ feature branch described in Equation \ref{eq:m}.

\section{Experiments}
In Section 4.1 we first present the public large-scale datasets utilized for CVGL along with the evaluation protocols. In Section 4.2, the implementation details of the model are thoroughly discussed. In Section 4.3, we provide a comparative analysis of the proposed method performance against state-of-the-art approaches. Finally, Section 4.4 presents the results of the ablation studies.

\subsection{Dataset and evaluation protocols}
\begin{figure}[h]
\centering
\includegraphics[width=\textwidth, keepaspectratio=true]{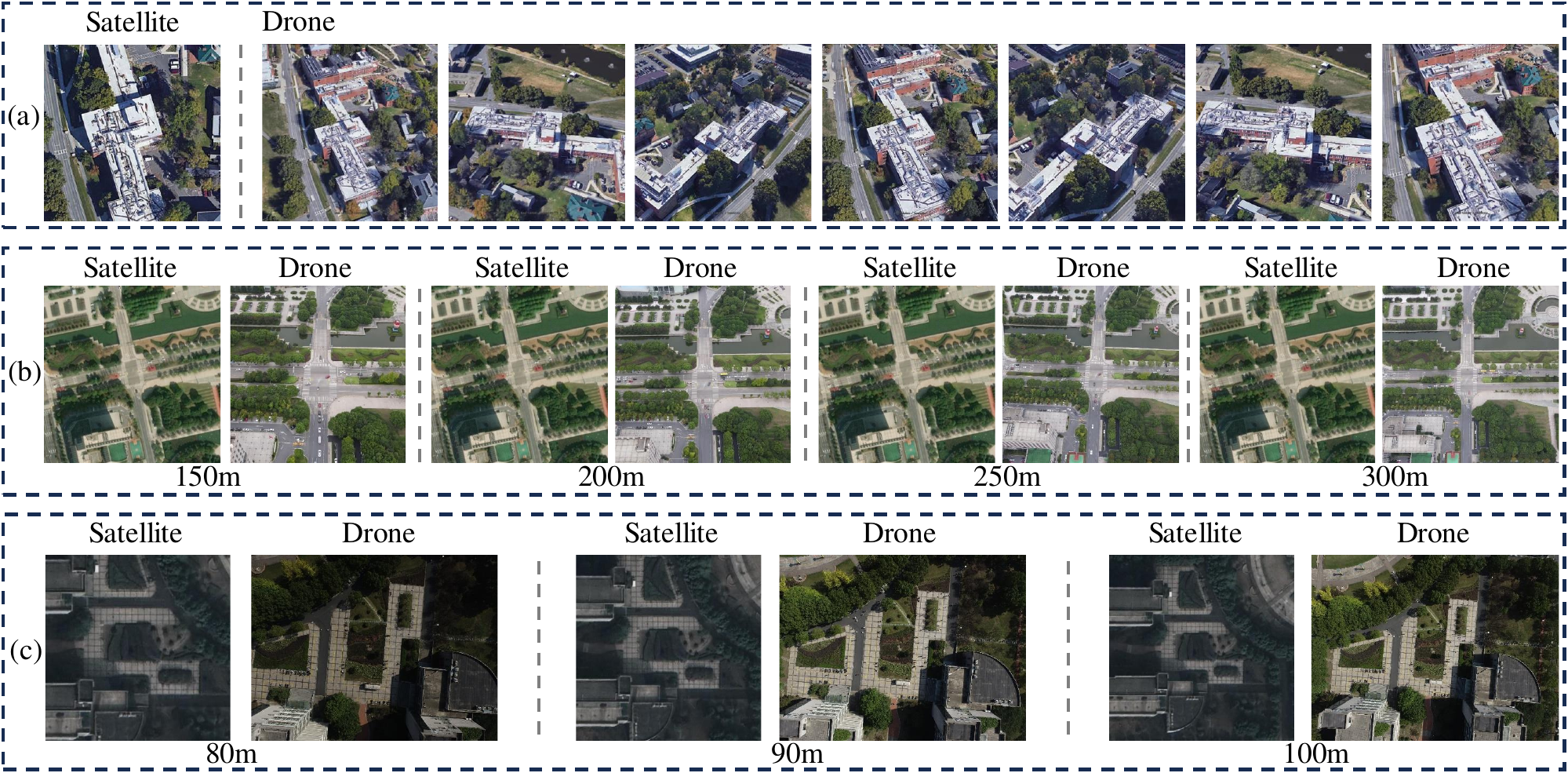}
\caption{Sample images of three public datasets. (a) row represent sample images of the same scene from different views in the University-1652 dataset. (b) row represent sample images of the same scene from various views at varying heights in the SUES-200 dataset. (c) row represent sample images of the same scene from different views at three heights in the Dense-UAV dataset.}
\label{fig:dataset}
\end{figure}

\textbf{(1) Unversity-1652 Dataset.} The University-1652 dataset comprises image data of 1,652 buildings across 72 universities worldwide. This dataset exhibits multi-view characteristics, with each location represented by one satellite image and 54 drone images. Figure \ref{fig:dataset}(a) presents sample images from different perspectives within the University-1652 dataset, while Table \ref{tab:dataset} provides a detailed distribution of images across categories, ensuring that the training and testing sets do not overlap. The dataset introduces two key tasks: drone localization (Drone→Satellite, D2S) and drone navigation (Satellite→Drone, S2D). During testing, in the D2S task, each drone-view query image corresponds to a single matched satellite-view image. The query set contains 37,855 drone images, and the gallery includes 701 genuine matching satellite images, alongside 250 satellite interference images. In the S2D task, the query set consists of 701 satellite images, and the gallery contains 37,855 corresponding drone images, along with 13,500 drone interference images.

\begin{table}[h]
\centering
\footnotesize
\caption{\centering The details of the three datasets, including the division of the training set and test set, as well as the number of images in the query set and the gallery set.}
\label{tab:dataset}
\def\tabblank{\hspace*{5mm}} 
\begin{tabularx}{\textwidth}
{@{\tabblank}@{\extracolsep{\fill}}c *{6}{c}@{\tabblank}} 
\toprule
\multirow{3}{*}{\textbf{Dataset}} & \multicolumn{2}{c}{\textbf{Training phase}} & \multicolumn{4}{c}{\textbf{Test phase}} \\ \cline{2-3} \cline{4-7} 
 & \multirow{2}{*}{Drone} & \multirow{2}{*}{Satellite} & \multicolumn{2}{c}{Drone→Satellite} & \multicolumn{2}{c}{Satellite→Drone} \\ \cline{4-5} \cline{6-7}
 &  &  & Query & Gallery & Query & Gallery \\ \toprule
University-1652\cite{U_1652} & 37854 & 701 & 37855 & 951 & 701 & 51355 \\ 
SUES-200\cite{SUES_200} & 24000 & 120 & 16000 & 200 & 50 & 40000 \\ 
Dense-UAV\cite{DenseUAV} & 6768 & 13536 & 2331 & 18198 & 4662 & 9099 \\ 
\bottomrule
\end{tabularx}
\end{table}

\textbf{(2) SUES-200 Dataset.} The SUES-200 dataset comprises both drone and satellite perspective images, following the same format as the University-1652 dataset. Contrasting with University-1652, the SUES-200 dataset includes four distinct altitudes: 150 meters, 200 meters, 250 meters and 300 meters. The SUES-200 dataset encompasses a broader variety of scene types, incorporating not only campus buildings but also parks, schools, lakes, and public buildings across 200 distinct locations. At each altitude, the dataset contains 50 drone-view images, while each location is represented by a single satellite-view image. SUES-200 is divided into training and testing sets, with 120 locations designated for training and 80 for testing. Figure \ref{fig:dataset}(b) illustrates sample images from the SUES-200 dataset, captured from varying perspectives and altitudes, and the detailed statistics of the dataset are provided in Table \ref{tab:dataset}.

\textbf{(3) Dense-UAV Dataset.} The Dense-UAV dataset differs from most existing datasets that rely on synthetic data from platforms, as it features drone-view images captured from real-world scenes at 14 universities in Zhejiang Province, China. The dataset includes images taken from three heights: 80m, 90m and 100m. As detailed in Table \ref{tab:dataset}, the training set consists of 6,768 drone-view images and 13,536 satellite-view images across 10 universities. The query set in the test set contains 2,331 drone-view images and 4,662 satellite-view images from 777 sampling points across 4 universities. The gallery set includes 27,297 images from 3,033 sampling points across all 14 universities. Figure \ref{fig:dataset}(c) presents image samples from the Dense-UAV dataset captured from different views and heights.

\textbf{(4) Evaluation protocols.} In image retrieval tasks, Recall@K (R@K) and Average Precision (AP) are commonly used to evaluate model performance. R@K denotes the proportion of correctly matched images within the top K ranked results. AP calculates the mean precision across different recall rates and is derived by computing the area under the precision-recall curve, thereby providing a comprehensive reflection of the balance between precision and recall. Furthermore, Spatial Distance Metric @K (SDM@K) serves as a composite evaluation metric in the Dense-UAV drone localization task. SDM@K integrates the advantages of Recall@K by assessing the spatial Euclidean distance between the query image and the top K most similar images in the gallery, thereby measuring localization accuracy. The range of this metric spans from 0 to 1, with values closer to 1 signifying superior localization performance.

\subsection{Implementation details}
The EVA02-L model utilizes pre-trained weights as the backbone for feature extraction. During training, the input images are resized to 448×448, and a series of data augmentation techniques are applied, including random padding, random rotation, random cropping, random flipping and random erasure. The optimization process employs Stochastic Gradient Descent (SGD) with a momentum of 0.9, a weight decay coefficient of 0.0005, and a batch size of 8. The learning rate for the backbone is set to 0.003, while the learning rate for the remaining layers is 0.01. The learning rate for all parameters is reduced by a factor of 0.1 at the 80th and 120th epochs, with a total training duration of 200 epochs. All experiments are conducted within the PyTorch, utilizing the NVIDIA A800 Tensor Core GPU.

\subsection{Comparison with State-of-the-art}
\begin{table}[h]
\footnotesize
\caption{Comparison with state-of-the-art methods on University-1652. The best results have been bolded.}
\label{tab:u1652 sota}
\tabcolsep 15pt  
\begin{tabular*}{0.9\textwidth}{cccccc}
\toprule
\multirow{2}{*}{\textbf{Method}} & \multirow{2}{*}{\textbf{Backbone}}& \multicolumn{2}{c}{\textbf{Drone→Satellite}} & \multicolumn{2}{c}{\textbf{Satellite→Drone}} \\
 \cmidrule(r{0.2cm}){3-4}
 \cmidrule(l{0.1cm}r{0.55cm}){5-6}
 &  & R@1 & AP & R@1 & AP \\
\midrule
University-1652\cite{U_1652} & ResNet-50 & 58.49 & 63.13 & 71.18 & 58.74 \\
LCM\cite{LCM} & ResNet-50 & 66.65 & 70.82 & 79.89 & 65.38 \\
LPN\cite{LPN} & ResNet-50 & 75.93 & 79.14 & 86.45 & 74.79 \\
PCL\cite{PCL} & ResNet-50 & 83.27 & 87.32 & 91.78 & 82.18 \\
FSRA\cite{FSRA} & ViT & 81.9 & 84.43 & 87.87 & 81.29 \\
Sample4Geo\cite{Deuser} & ConvNeXt & 92.65 & 93.81 & 95.14 & 91.39 \\
TransFG\cite{Transfg} & ViT & 84.01 & 86.31 & 90.16 & 84.61 \\
MCCG\cite{MCCG} & ConvNeXt & 89.64 & 91.32 & 94.58 & 89.85 \\
SDPL\cite{SDPL} & SwinV2 & 90.16 & 91.64 & 93.58 & 89.45 \\
Ours& EVA02 & \textbf{95.06} & \textbf{95.89} & \textbf{96.01} & \textbf{95.07} \\
\bottomrule
\end{tabular*}
\end{table}

\begin{figure}[!t]
\centering
\includegraphics[width=\textwidth, keepaspectratio=true]{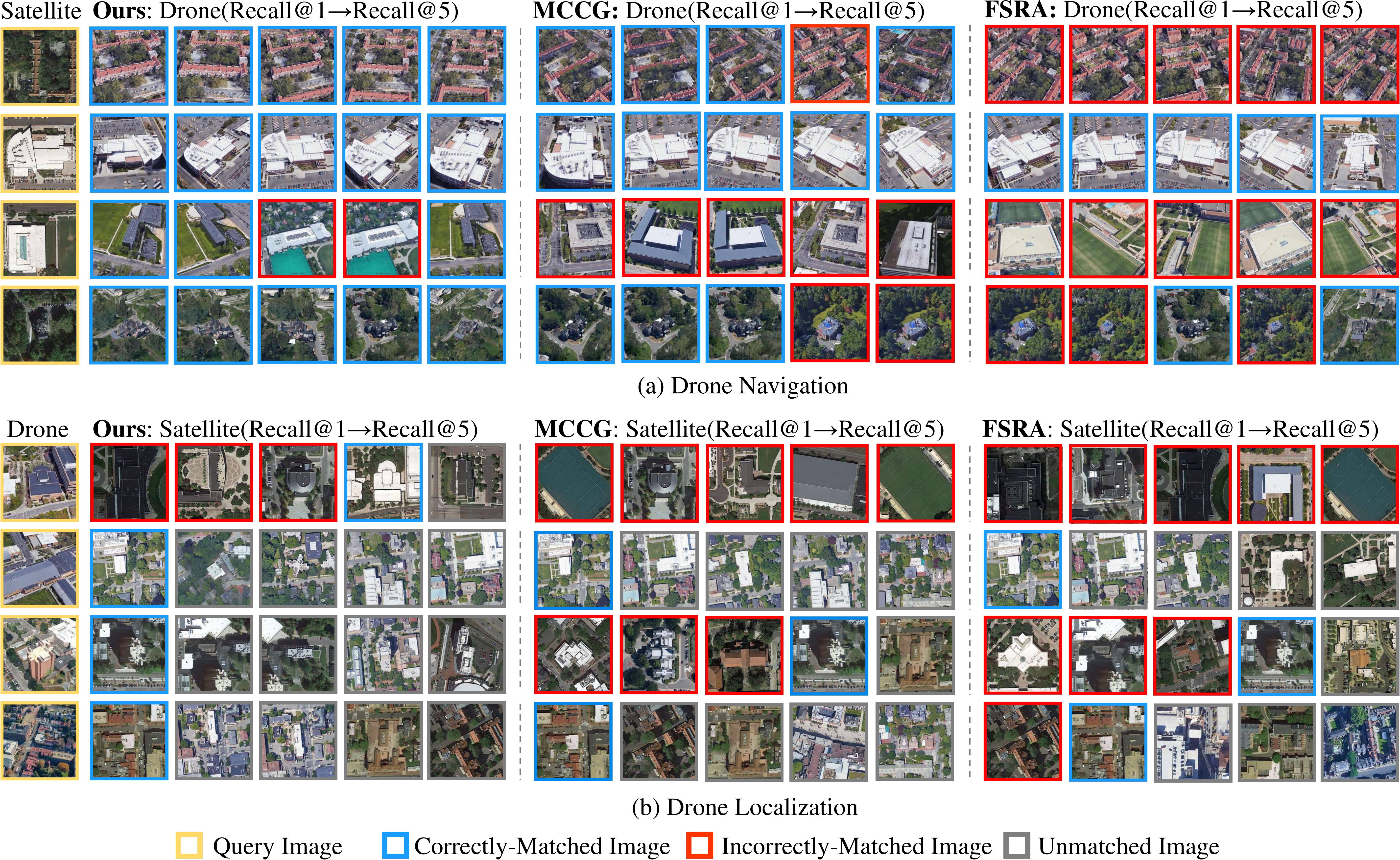}
\caption{Image retrieval results obtained with Ours, MCCG and FSRA. (a) Top-5 retrieval results of drone navigation on University-1652; (b) Top-5 retrieval results of drone localization on University-1652.}
\label{fig:visualization}
\end{figure}
\textbf{(1) Results on University-1652.} As shown in Table \ref{tab:u1652 sota}, in the D2S task, the MFAF method achieves a R@1 of 95.06$\%$ and an AP of 95.89$\%$, surpassing MCCG by 5.52$\%$ and 4.57$\%$, and FSRA by 13.16$\%$ and 11.46$\%$ respectively. In the S2D task, the MFAF method obtains 96.01$\%$ for R@1 and 95.07$\%$ for AP, exceeding the MCCG by 1.43$\%$ and 5.22$\%$, and FSRA by 8.14$\%$ and 13.78$\%$. And the Figure \ref{fig:visualization} illustrates representative retrieval results on both tasks. In drone navigation, four satellite-view images are randomly selected as queries, from which the model retrieves the top five most similar drone-view images from the gallery. Conversely, in drone localization, four drone-view images are used as queries, each corresponding to a single ground-truth satellite-view match within the gallery. The top five most similar satellite images are retrieved, with correctly matched results highlighted, while unmatched results are indicated with gray borders. When query images contain clear and distinctive targets (e.g., the second row in Figure \ref{fig:visualization} (a) and Figure \ref{fig:visualization} (b)), all three methods demonstrate successful retrieval. However, in scenarios where the architectural targets are blurry or confounded by visually similar backgrounds (e.g., the first and fourth rows in Figure \ref{fig:visualization} (a), and the third and fourth rows in Figure \ref{fig:visualization} (b)), MCCG and FSRA exhibit retrieval errors, while MFAF consistently identifies the correct matches, demonstrating its superior ability to distinguish visually similar content. This advantage is attributed to the proposed multi-frequency feature fusion strategy, which simultaneously preserves global representations and captures fine-grained details, enhancing cross-view feature discriminability. Compared to conventional CNN- and ViT-based methods, which often focus on central features or overlook global consistency during feature segmentation, the proposed MFAF method makes breakthroughs in several aspects by incorporating global feature attention and effectively extracting relevant frequency information from the image. Although a slight performance drop occurs in certain challenging cases (e.g., the third row in Figure \ref{fig:visualization} (a) and first row in Figure \ref{fig:visualization} (b)), competing methods fail entirely under such extreme viewpoint discrepancies, further highlighting MFAF’s robustness. 

\begin{table}[h]
\tabcolsep 1.5pt
\footnotesize 
\caption{Comparison with state-of-the-art results on SUES-200.}
\label{tab:sues200 sota}
\begin{tabularx}{\textwidth}{@{}c *{16}{c}@{}} 
\toprule
\multirow{3}{*}{\textbf{Method}} & \multicolumn{8}{c}{\textbf{Drone→Satellite}} & \multicolumn{8}{c}{\textbf{Satellite→Drone}} \\ \cmidrule(r{0.2cm}){2-9} \cmidrule(l{0.1cm}){10-17}
 & \multicolumn{2}{c}{150m} & \multicolumn{2}{c}{200m} & \multicolumn{2}{c}{250m} & \multicolumn{2}{c}{300m} & \multicolumn{2}{c}{150m} & \multicolumn{2}{c}{200m} & \multicolumn{2}{c}{250m} & \multicolumn{2}{c}{300m} \\ 
 \cmidrule(r{0.2cm}){2-9}
 \cmidrule(l{0.1cm}){10-17}
 & R@1 & AP & R@1 & AP & R@1 & AP & R@1 & AP & R@1 & AP & R@1 & AP & R@1 & AP & R@1 & AP \\ \toprule
SUES-200\cite{SUES_200} & 59.32 & 64.93 & 62.30 & 67.24 & 82.50 & 58.95 & 85.00 & 62.56 & 82.50 & 58.95 & 85.00 & 62.56 & 88.75 & 69.96 & 96.25 & 84.16 \\
FSRA\cite{FSRA} & 68.25 & 73.45 & 83.00 & 85.99 & 83.75 & 76.67 & 90.00 & 85.34 & 83.75 & 76.67 & 90.00 & 85.34 & 93.75 & 90.17 & 95.00 & 92.03 \\
MCCG\cite{MCCG} & 82.22 & 85.47 & 89.38 & 91.41 & 93.75 & 89.72 & 93.75 & 92.21 & 93.75 & 89.72 & 93.75 & 92.21 & 96.25 & 96.14 & 98.75 & 96.64 \\
SDPL\cite{SDPL} & 82.95 & 85.82 & 92.73 & 94.07 & 93.75 & 83.75 & 96.25 & 92.42 & 93.75 & 83.75 & 96.25 & 92.42 & 97.50 & 95.65 & 96.25 & 96.17 \\
Ours & \textbf{95.22} & \textbf{96.30} & \textbf{98.80} & \textbf{99.08} & \textbf{97.50} & \textbf{96.87} & \textbf{97.50} & \textbf{98.42} & \textbf{97.50} & \textbf{96.87} & \textbf{97.50} & \textbf{98.03} & \textbf{98.75} & \textbf{97.98} & \textbf{98.75} & \textbf{98.72} \\ \bottomrule
\end{tabularx}
\end{table}

\textbf{(2) Results on SUES-200.} Table \ref{tab:sues200 sota} presents a comparative analysis between MFAF and existing state-of-the-art methods, demonstrating the superior performance of MFAF across various metrics. In the D2S task, compared to the SDPL method, MFAF improves the R@1 by 12.27$\%$, 6.07$\%$, 2.33$\%$ and 0.70$\%$ at four heights, while the AP increases by 10.48$\%$, 5.01$\%$, 2.02$\%$ and 0.84$\%$. In comparison with the MCCG method, MFAF enhances R@1 by 13.00$\%$, 9.42$\%$, 4.56$\%$ and 3.46$\%$, and boosts AP by 10.83$\%$, 7.67$\%$, 3.67$\%$ and 2.69$\%$ at four heights. For the S2D task, MFAF achieves R@1 of 97.50$\%$, 97.50$\%$, 98.75$\%$ and 98.75$\%$ across the four heights, with corresponding AP of 96.87$\%$, 98.03$\%$, 97.98$\%$ and 98.72$\%$, consistently delivering optimal performance. These results underscore the excellent and stable performance of MFAF across different heights.

\begin{table}[h]
\centering
\footnotesize
\caption{Comparison with state-of-the-art results on Dense-UAV.}
\label{tab:Dense sota}
\def\tabblank{\hspace*{5mm}} 
\begin{tabularx}{0.6\textwidth}
{@{\tabblank}@{\extracolsep{\fill}}c *{3}{c}@{\tabblank}}
\toprule
\textbf{Method} & \textbf{R@1} & \textbf{R@5} & \textbf{SDM@1} \\ \hline
DenseUAV\cite{DenseUAV} & 83.01 & 95.58 & 86.50 \\
FSRA\cite{FSRA}     & 81.21 & 94.55 & 85.11 \\
MCCG\cite{MCCG}     & 89.19 & 96.87 & 90.99 \\
Ours & \textbf{95.22} & \textbf{99.93} & \textbf{95.23} \\ \bottomrule
\end{tabularx}
\end{table}

\textbf{(3) Results on Dense-UAV.} We further validated the effectiveness of the MFAF method on the Dense-UAV dataset, with the detailed results provided in Table \ref{tab:Dense sota}. In the D2S task, MFAF achieved optimal performance with R@1, R@5 and SDM@1 of 95.22$\%$, 99.93$\%$ and 95.23$\%$ respectively. Compared to the MCCG method, MFAF improved R@1, R@5 and SDM@1 by 6.03$\%$, 3.06$\%$, and 4.24$\%$ respectively.

\subsection{Ablation study}

\begin{table}[h]
\footnotesize
\caption{Results of ablation experiments of the proposed MFAF on University-1652.}
\label{tab:MAFA}
\tabcolsep 2.8pt
\def\tabblank{\hspace*{0mm}} 
\begin{tabularx}{\textwidth}
{@{\tabblank}@{\extracolsep{\fill}}c *{4}{c} c *{4}{c}@{\tabblank}}
\toprule
\multirow{2}{*}{\textbf{Method}} & \multicolumn{2}{c}{\textbf{Drone→Satellite}} & \multicolumn{2}{c}{\textbf{Satellite→Drone}} & \multirow{2}{*}{\textbf{Method}} & \multicolumn{2}{c}{\textbf{Drone→Satellite}} & \multicolumn{2}{c}{\textbf{Satellite→Drone}} \\ \cline{2-3} \cline{4-5} \cline{7-8} \cline{9-10} 
 & R@1 & AP & R@1 & AP & & R@1 & AP & R@1 & AP \\ \toprule
Baseline(EVA02-L) & 75.96 & 79.08 & 84.31 & 77.54 & Baseline(SwinV2-L) & 73.95 & 77.23 & 84.02 & 74.20 \\
+MFB & 94.29 & 95.24 & 95.44 & 94.64 & +MFB & 91.74 & 92.87 & 94.15 & 90.96 \\
+FSA & \textbf{95.06} & \textbf{95.89} & \textbf{96.01} & \textbf{95.07} & +FSA & \textbf{92.26} & \textbf{93.49} & \textbf{94.72} & \textbf{91.61} \\ \bottomrule
\end{tabularx}
\end{table}

\textbf{(1) Ablation experiments of the proposed method.} Table \ref{tab:MAFA} presents the ablation study results for the proposed MFAF method, where baseline models are SwinV2-L and EVA02-L. The objective of this experiment is to analyze the contributions of the MFB and FSA in the MFAF framework. For both baseline models, the MFB captures the global structural similarities (such as the overall layout of buildings and the general contours of the terrain) of images from different views, and further enhances the model ability to express image details (e.g. the edge shapes of buildings and the texture direction of roads) by fusing multi-scale frequency features. This improvement strengthens the discrimination of the features, significantly increasing the cross-view accuracy. In the D2S task with EVA02-L as the baseline, the MFB led to an 18.33$\%$ and 16.16$\%$ increase in R@1 and AP. In the S2D task, R@1 and AP were improved by 11.13$\%$ and 17.1$\%$. Similarly, for the D2S task with SwinV2-L as the baseline, the addition of MFB resulted in a 17.79$\%$ and 15.64$\%$ increase in R@1 and AP; in the S2D task, R@1 and AP increased by 10.13$\%$ and 16.76$\%$. Furthermore, the FSA module introduces frequency-awareness, which enhances the attention computation to focus on key regions of global and detailed features by combining the spatial distribution features of low-frequency global structures and high-frequency detailed textures. This strategy effectively reduces interference from background noise and viewpoint variations, and enhances the cross-view robustness. After incorporating FSA, the D2S task achieved an R@1 of 95.06$\%$ and an AP of 95.89$\%$, while the S2D task attained an R@1 of 96.01$\%$ and an AP of 95.07$\%$. Furthermore, the MFAF (SwinV2-L) with the addition of FSA, exhibited further improvements, with R@1 and AP increasing by 0.52$\%$ and 0.62$\%$ in D2S task; in S2D task, R@1 and AP were improved by 0.57$\%$ and 0.65$\%$, demonstrating superior performance compared to the SDPL method.

\begin{table}[h]
\centering
\footnotesize
\caption{Results of different pooling strategies in FSA.}
\label{tab:pooling}
\def\tabblank{\hspace*{5mm}} 
\begin{tabularx}{0.6\textwidth}
{@{\tabblank}@{\extracolsep{\fill}}c *{4}{c}@{\tabblank}} 
\toprule
\multirow{2}{*}{\textbf{Pooling}} & \multicolumn{2}{c}{\textbf{Drone→Satellite}} & \multicolumn{2}{c}{\textbf{Satellite→Drone}} \\  \cmidrule(r{0.2cm}){2-3}
 \cmidrule(l{0.1cm}){4-5}
 & R@1 & AP & R@1 & AP \\ \toprule
Z-Pool & \textbf{95.06} & \textbf{95.89} & \textbf{96.01} & \textbf{95.07} \\
AAP    & 93.95 & 94.96 & 95.72 & 94.15 \\
AP     & 94.96 & 95.74 & 95.86 & 94.93 \\
MP     & 82.78 & 84.92 & 87.45 & 82.95 \\ \bottomrule
\end{tabularx}
\end{table}
\textbf{(2) Effect of pooling strategies in FSA.} Table \ref{tab:pooling} presents a comparison of four pooling methods, including Z-Pooling(Z-Pool), Adaptive Average Pooling(AAP), Average Pooling(AP) and Max Pooling(MP). The experimental results indicate that the Z-Pooling, which combines average pooling and max pooling, yields the most optimal performance.
\begin{table}[h]
\footnotesize
\centering
\caption{Result of different branch strategies in MFAF}
\label{tab:branch}
\def\tabblank{\hspace*{5mm}} 
\begin{tabularx}{0.8\textwidth}
{@{\tabblank}@{\extracolsep{\fill}}c c c *{4}{c}@{\tabblank}} 
\toprule
\multicolumn{3}{c}{\textbf{Method}} & \multicolumn{2}{c}{\textbf{Drone→Satellite}} & \multicolumn{2}{c}{\textbf{Satellite→Drone}} \\ \cline{1-3} \cline{4-5} \cline{6-7} 
\text{HF branch} & LF branch & Original branch & R@1 & AP & R@1 & AP \\ \toprule
\checkmark &  & \checkmark & 92.95 & 94.05 & 94.72 & 93.45 \\
 & \checkmark & \checkmark & 94.55 & 95.43 & 96.15 & 94.27 \\
\checkmark & \checkmark & \checkmark & 95.06 & 95.89 & 96.01 & 95.07 \\ \bottomrule
\end{tabularx}
\end{table}
\begin{figure}[h]
\centering
\includegraphics[width=\textwidth, keepaspectratio=true]{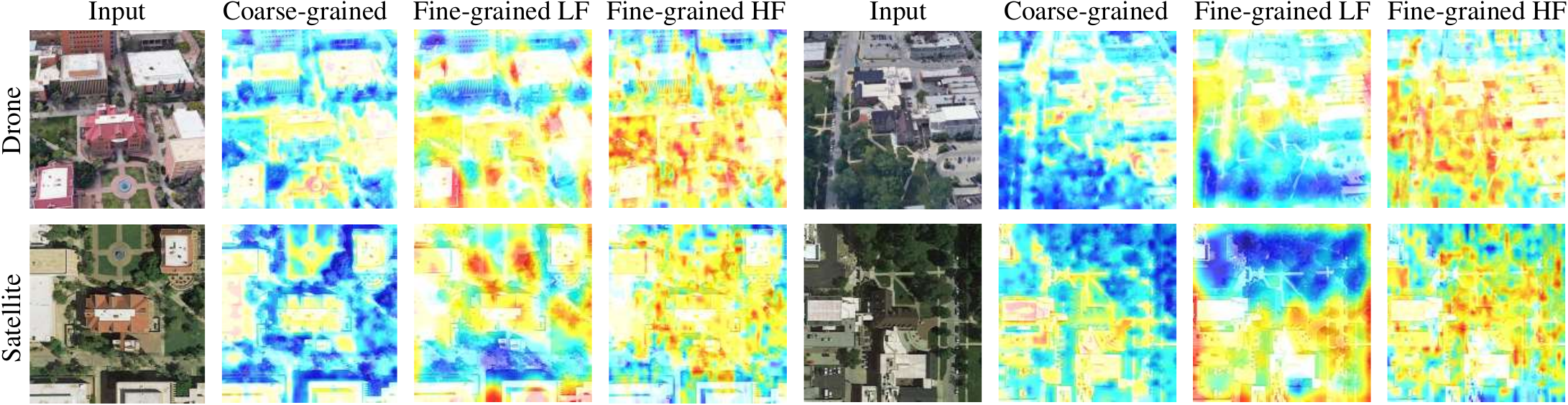}
\caption{Heatmaps generated by coarse-grained and fine-grained branches of MFAF for drone and satellite views.}
\label{fig：热力图-高低频}
\end{figure}

\textbf{(3) Effect of each branch in MFAF.} This experiment aims to assess the contributions of high-frequency, low-frequency and original branches within the model. When the high-frequency branch was removed, compared to the complete model, the R@1 and AP in the D2S task decreased by 2.11$\%$ and 1.84$\%$; in the S2D task, R@1 and AP dropped by 1.29$\%$ and 1.62$\%$ as shown in Table \ref{tab:branch}. This can be attributed to the high-frequency branch ability to provide fine-grained detail features distinct from those of the low-frequency and original branches. In contrast, the removal of the low-frequency branch resulted in a smaller decline in R@1 and AP across both tasks. This is because the low-frequency branch extracts global structural features, which effectively compensate for the high-frequency branch limitations in global information processing, thereby enhancing the model global matching capacity in cross-view and improving positioning accuracy. Additionally, the original branch preserves unprocessed, comprehensive features, maintaining the model’s foundational integrity and compensating for any information loss during feature extraction in the high-frequency and low-frequency branches. Figure \ref{fig：热力图-高低频} shows heatmaps of the global coarse-grained features from the original branch generated by EVA02 as the backbone and the fine-grained features from the low-frequency (LF) and high-frequency (HF) branches produced by the MFAF module. The global coarse-grained branch focuses on overall architectural information, while the LF branch enhances the extraction of global features, and the HF branch captures fine details such as roads, trees and building edges. This fusion of features, combining original and frequency-processed information, enriches the feature representation, significantly improving the model’s discriminative power and consistency in CVGL.

\begin{figure}[h]
\centering
\includegraphics[width=\textwidth, keepaspectratio=true]{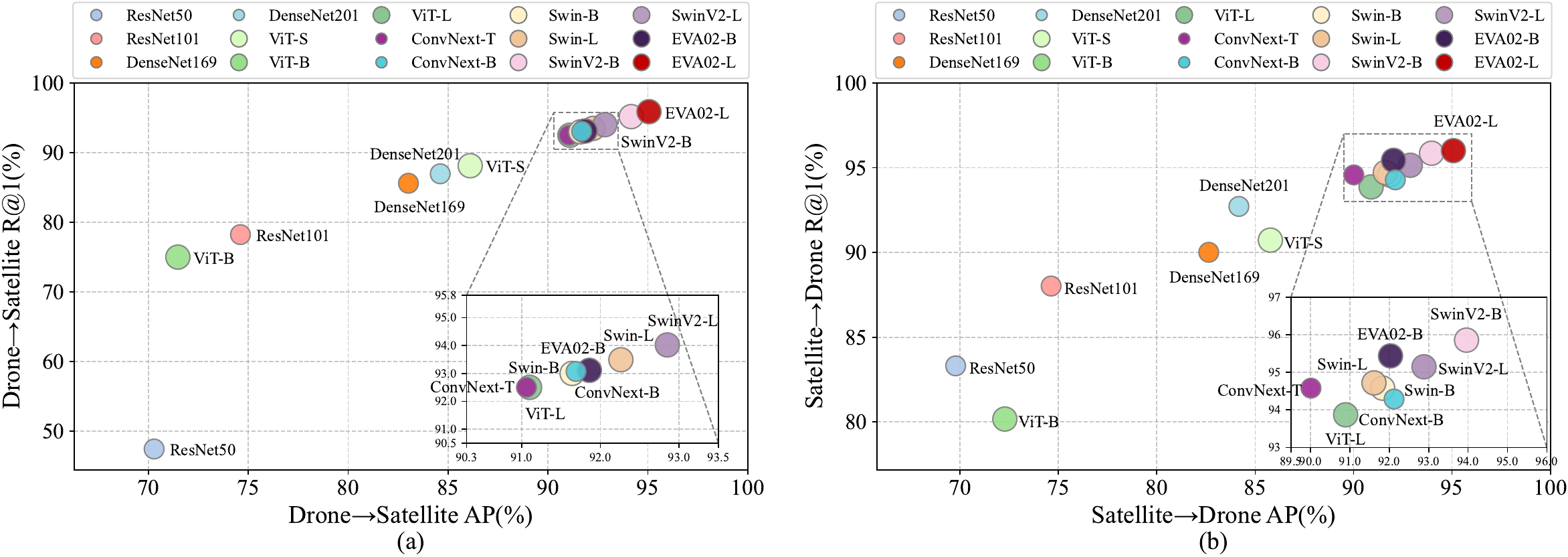}
\caption{Comparison for different backbones. (a) Comparison results on the Drone→Satellite task; (b) Comparison results on the Satellite→Drone task. The horizontal axis represents AP accuracy, and the vertical axis represents R@1 accuracy. Large circles indicate Transformer-based backbones, while small circles indicate CNN-based backbones. }
\label{fig:backbone}
\end{figure}

\textbf{(4) Effect of EVA02 in cross-view.} To evaluate the effectiveness of the EVA02, we conducted a comparative analysis with several prevalent backbones. In the experiment, we selected ResNet, DenseNet, ViT, ConvNeXt, Swin Transformer and Swin Transformer V2, with detailed results presented in Figure \ref{fig:backbone}. For both CVGL tasks, the EVA02-L achieved the best performance in terms of R@1 and AP metrics. Additionally, we compared the accuracy of two backbone with different depths, EVA02-B and EVA02-L. The results demonstrated that EVA02-L outperforms EVA02-B, which can be attributed to the performance gains arising from its more intricate main chain structure.

\begin{table}[h]
\centering
\footnotesize
\caption{Result of different input image sizes.}
\label{tab:imagesize}
\centering
\tabcolsep 10.1pt  
\begin{tabular*}{\textwidth}{cccccccc}
\toprule
\multirow{2}{*}{\textbf{Image Size}} 
& \multirow{2}{*}{\textbf{Training(mins)}} 
& \multicolumn{3}{c}{\textbf{Drone→Satellite}} 
& \multicolumn{3}{c}{\textbf{Satellite→Drone}} \\ 
\cmidrule(l{0cm} r{0cm}){3-5}
\cmidrule(l{0.3cm} r{0cm}){6-8}
& & Inference(mins) & R@1 & AP & Inference(mins) & R@1 & AP \\
\midrule
$196 \times 196$ & 237.57 & 17.33 & 88.05 & 89.79 & 23.65 & 92.01 & 87.95 \\ 
$280 \times 280$ & 251.23 & 29.73 & 93.92 & 95.06 & 40.45 & 95.58 & 94.29 \\ 
$378 \times 378$ & 259.30 & 50.13 & 94.44 & 95.33 & 65.20 & 96.01 & 94.78 \\ 
\textbf{\boldmath $448 \times 448$} & 294.20 & 69.15 & \textbf{95.06} & \textbf{95.89} & 100.82 & 96.01 & \textbf{95.07} \\ 
$588 \times 588$ & 449.33 & 123.73 & 94.84 & 95.75 & 166.17 & \textbf{96.15} & 95.03  \\
\bottomrule
\end{tabular*}
\end{table}
\textbf{(5) Effect of the input image size.} As shown in Table \ref{tab:imagesize}, the model performance improves progressively with the increase in image resolution, reaching optimal performance at a resolution of 448 pixels, after which a decline is observed at 588 pixels. Consequently, 448 pixels was selected as the input image size for the MFAF method. Note that larger image sizes not only demand more memory but also result in longer inference time, while smaller image sizes may compress image information, leading to a diminished ability. In scenarios with limited computational resources, smaller image sizes may be preferable

\begin{figure}[h]
\centering
\includegraphics[width=0.7\textwidth, keepaspectratio=true]{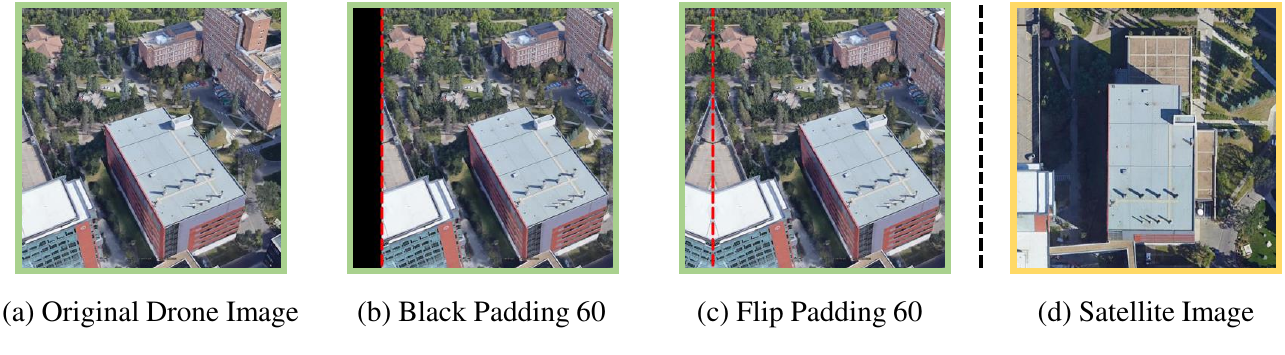}
\caption{The images of Black-Padding 60 and Flip-Padding 60. (a) Original drone-view image; (b) BP image, where a 60-pixel-wide black strip is inserted on the left side, while an equivalent 60-pixel portion is removed from the right side; (c) FP image, where a 60-pixel-wide section from the left side is mirrored and appended to the left, with a corresponding 60-pixel portion removed from the right side; (d) Corresponding satellite-view image. The red dashed line is the boundary for the padding operation.}
\label{fig:padding}
\vspace{-\baselineskip}
\end{figure}
\begin{table}[h]
\footnotesize
\caption{Effect of different pad sizes on the rate of decline of AP and R@1 values for different methods.}
\label{tab:pad_shift}
\centering
\newlength{\padvalheight}
\newcommand{\padValue}[2]{%
  \settowidth{\padvalheight}{#1}
  \makebox[\padvalheight]{#1}
  -\makebox[0pt][l]{\scalebox{0.75}{#2}}
}

\newcolumntype{C}[1]{>{\centering\arraybackslash}p{#1}}
\begin{tabularx}{\textwidth}{%
  C{0.3cm}  
  *{12}{C{0.83cm}}  
}
\toprule
\textbf{Pixel}
 & \multicolumn{3}{c}{\textbf{BP AP(\%)}} 
 & \multicolumn{3}{c}{\textbf{FP AP(\%)}} 
 & \multicolumn{3}{c}{\textbf{BP Recall@1(\%)}} 
 & \multicolumn{3}{c}{\textbf{FP Recall@1(\%)}} \\
\cmidrule(l{0.3cm} r{0cm}){2-4}    
\cmidrule(l{0.1cm} r{0cm}){5-7}  
\cmidrule(l{0.1cm} r{0cm}){8-10}
\cmidrule(l{0.1cm} r{-0.2cm}){11-13}
\textbf{Num}
 & \textbf{ Ours} & \textbf{\,MCCG} & \textbf{\,FSRA} 
 & \textbf{ Ours} & \textbf{\,MCCG} & \textbf{\,FSRA} 
 & \textbf{ Ours} & \textbf{\,MCCG} & \textbf{\,FSRA} 
 & \textbf{ Ours} & \textbf{\,MCCG} & \textbf{\,FSRA} \\
\midrule
0  & \padValue{\textbf{95.89}}{\textbf{0.00}}   & \padValue{91.32}{0.00}   & \padValue{84.43}{0.00} 
    & \padValue{\textbf{95.89}}{\textbf{0.00}}  & \padValue{91.32}{0.00}   & \padValue{84.43}{0.00} 
    & \padValue{\textbf{95.06}}{\textbf{0.00}}  & \padValue{89.64}{0.00}   & \padValue{81.90}{0.00} 
    & \padValue{\textbf{95.06}}{\textbf{0.00}}  & \padValue{89.64}{0.00}   & \padValue{81.90}{0.00} \\
10 & \padValue{\textbf{95.76}}{\textbf{0.13}} & \padValue{89.84}{1.48} & \padValue{83.00}{1.43} 
    & \padValue{\textbf{95.73}}{\textbf{0.16}} & \padValue{89.35}{1.97} & \padValue{83.66}{0.77} 
    & \padValue{\textbf{94.90}}{\textbf{0.16}} & \padValue{88.03}{1.61} & \padValue{80.10}{1.80} 
    & \padValue{\textbf{94.86}}{\textbf{0.20}} & \padValue{87.43}{2.21} & \padValue{80.90}{1.00} \\
20 & \padValue{\textbf{95.63}}{\textbf{0.26}} & \padValue{89.21}{2.11} & \padValue{81.24}{3.19} 
    & \padValue{\textbf{95.41}}{\textbf{0.48}} & \padValue{87.08}{4.24} & \padValue{81.51}{2.92} 
    & \padValue{\textbf{94.74}}{\textbf{0.32}} & \padValue{87.28}{2.36} & \padValue{78.06}{3.84} 
    & \padValue{\textbf{94.46}}{\textbf{0.60}} & \padValue{84.74}{4.90} & \padValue{78.37}{3.85} \\
30 & \padValue{\textbf{95.35}}{\textbf{0.54}} & \padValue{87.90}{3.42} & \padValue{77.80}{6.63} 
    & \padValue{\textbf{94.88}}{\textbf{1.01}} & \padValue{82.94}{8.38} & \padValue{77.39}{7.04} 
    & \padValue{\textbf{94.37}}{\textbf{0.69}} & \padValue{85.72}{3.92} & \padValue{74.10}{7.80} 
    & \padValue{\textbf{93.78}}{\textbf{1.27}} & \padValue{79.89}{9.75} & \padValue{73.66}{8.24} \\
40 & \padValue{\textbf{95.00}}{\textbf{0.89}} & \padValue{85.53}{5.79} & \padValue{73.10}{11.3} 
    & \padValue{\textbf{94.03}}{\textbf{1.86}} & \padValue{76.84}{14.5} & \padValue{71.53}{12.9} 
    & \padValue{\textbf{93.94}}{\textbf{1.12}} & \padValue{82.91}{6.73} & \padValue{68.83}{13.1} 
    & \padValue{\textbf{92.71}}{\textbf{2.35}} & \padValue{72.94}{16.7} & \padValue{67.11}{14.8} \\
50 & \padValue{\textbf{94.34}}{\textbf{1.55}} & \padValue{82.37}{8.95} & \padValue{67.29}{17.1} 
    & \padValue{\textbf{92.78}}{\textbf{3.10}} & \padValue{68.64}{22.7} & \padValue{64.41}{20.0} 
    & \padValue{\textbf{93.10}}{\textbf{1.96}} & \padValue{79.24}{10.4} & \padValue{62.43}{19.5} 
    & \padValue{\textbf{91.15}}{\textbf{3.91}} & \padValue{63.86}{25.8} & \padValue{59.34}{22.6} \\
60 & \padValue{\textbf{93.33}}{\textbf{2.56}} & \padValue{77.71}{13.6} & \padValue{60.89}{23.5} 
    & \padValue{\textbf{90.81}}{\textbf{5.08}} & \padValue{59.47}{31.9} & \padValue{56.02}{28.4} 
    & \padValue{\textbf{91.82}}{\textbf{3.24}} & \padValue{73.91}{15.7} & \padValue{55.54}{26.4} 
    & \padValue{\textbf{88.69}}{\textbf{6.37}} & \padValue{54.08}{35.6} & \padValue{50.49}{31.4} \\
\bottomrule
\end{tabularx}
\end{table}
\begin{figure}[!h]
\vspace{-\baselineskip}
\centering
\begin{minipage}[c]{0.245\textwidth}
\centering
\includegraphics[width=\textwidth]{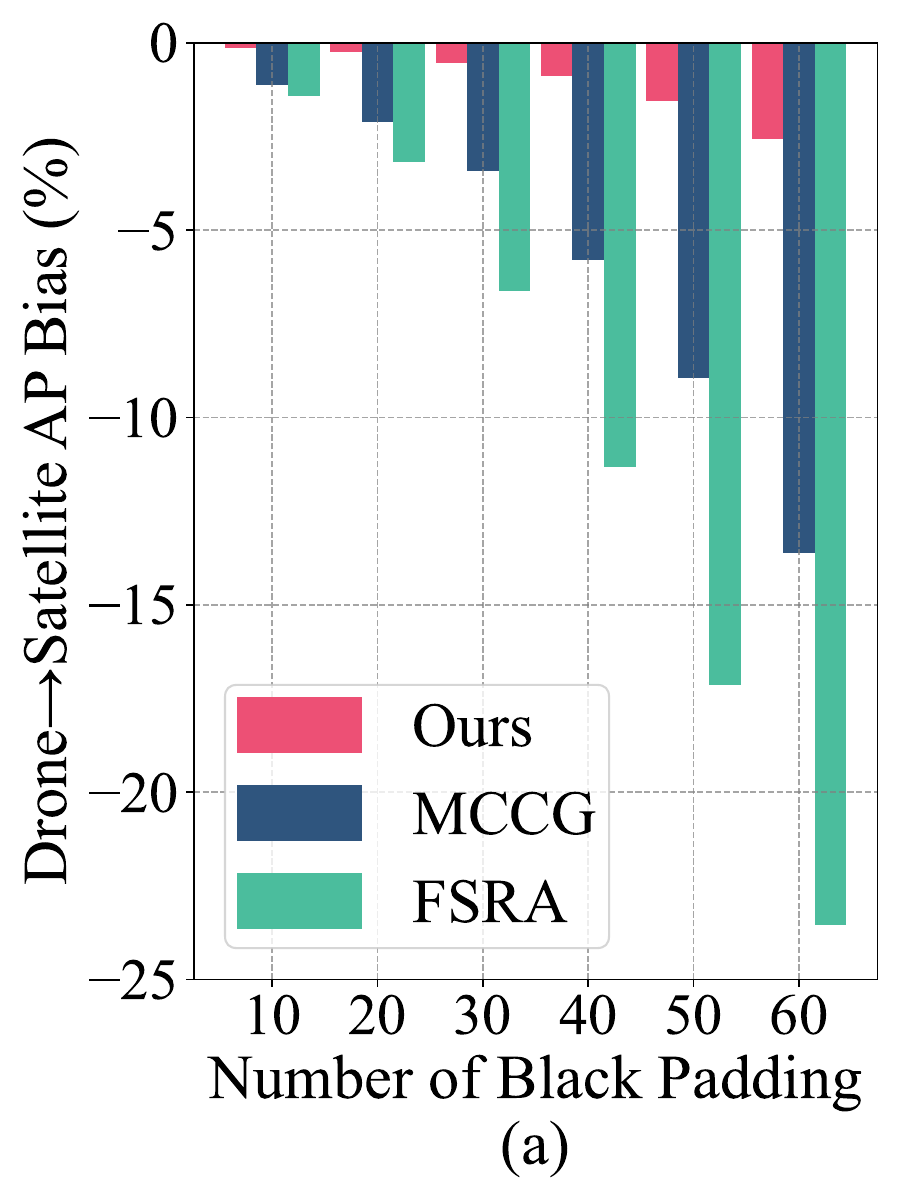}
\end{minipage}
\begin{minipage}[c]{0.245\textwidth}
\centering
\includegraphics[width=\textwidth]{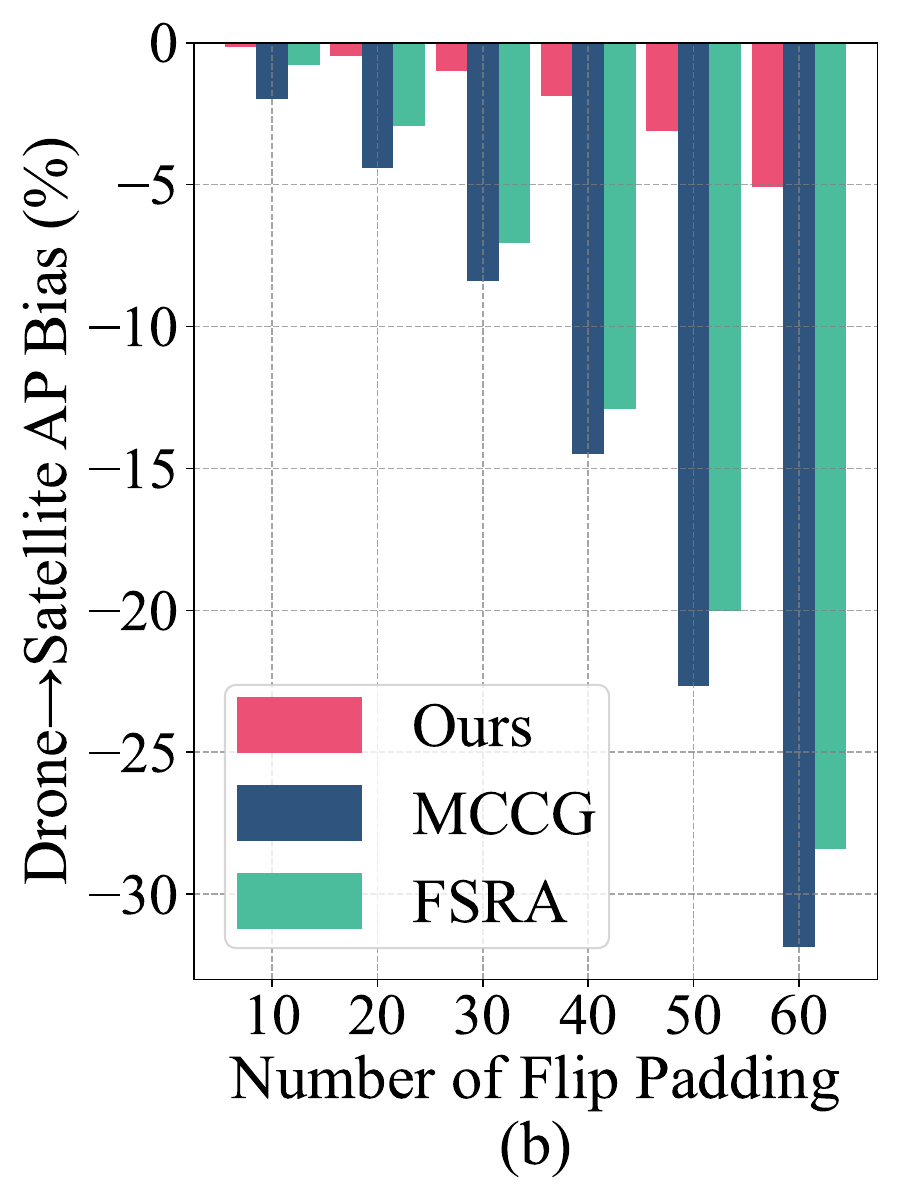}
\end{minipage}
\begin{minipage}[c]{0.245\textwidth}
\centering
\includegraphics[width=\textwidth]{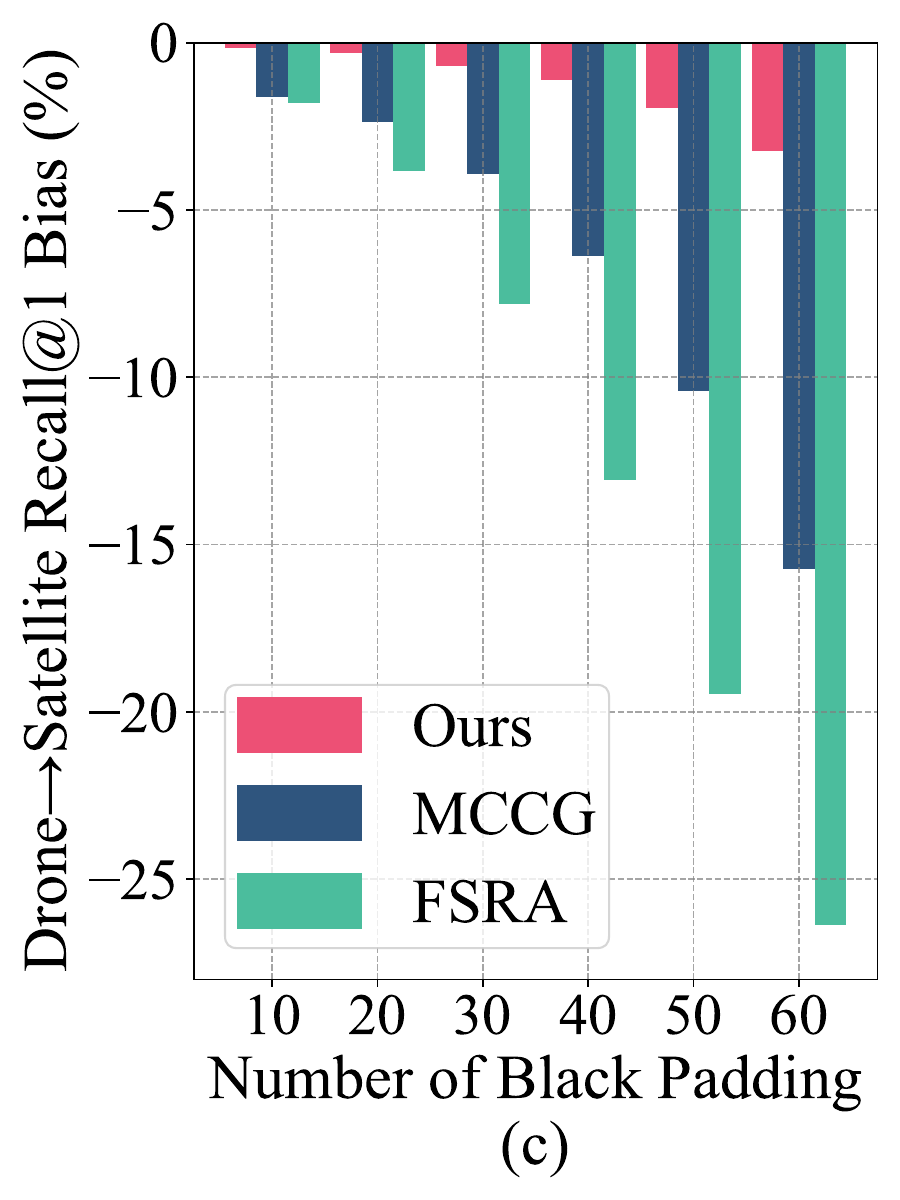}
\end{minipage}
\begin{minipage}[c]{0.245\textwidth}
\centering
\includegraphics[width=\textwidth]{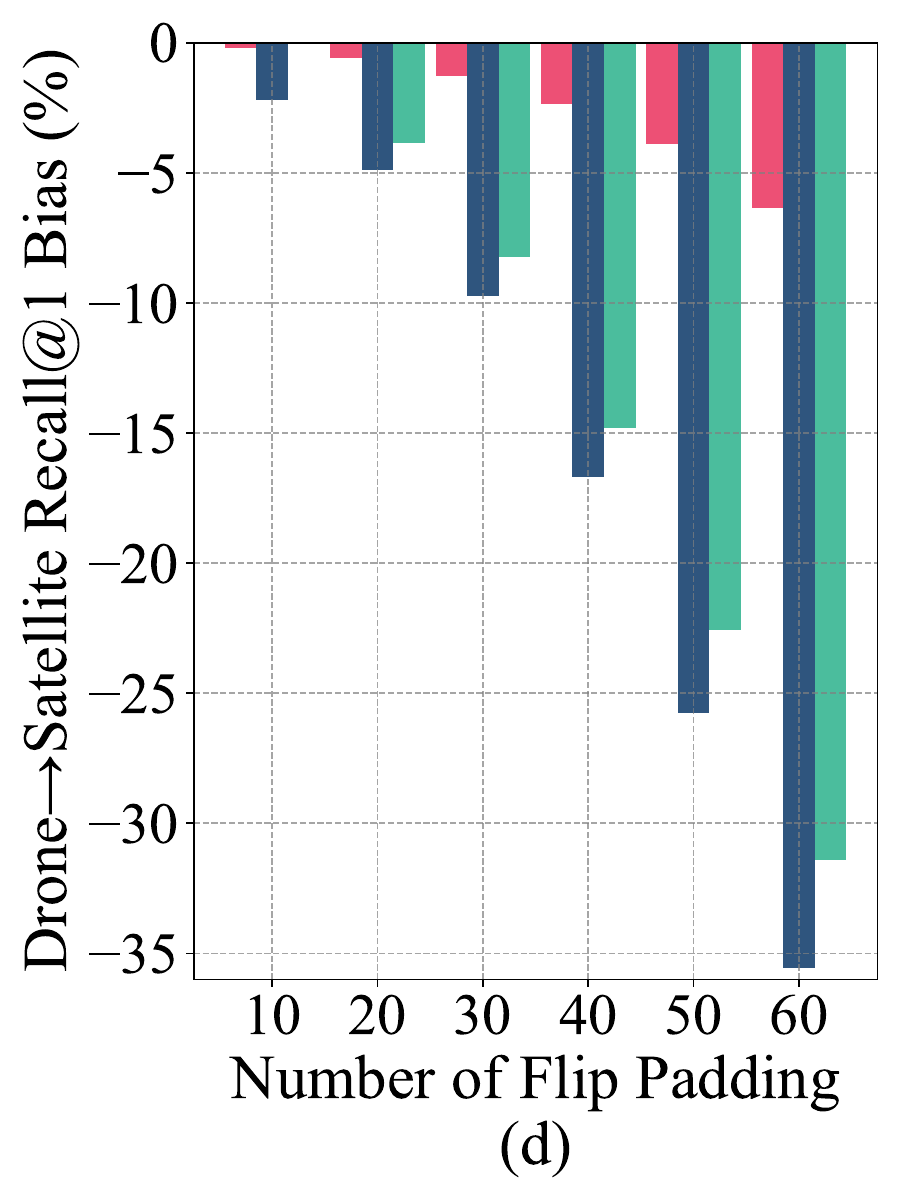}
\end{minipage}
\caption{Comparison of robustness to position shifting on MFAF(Ours), MCCG and FSRA. (a) Effect of Black Padding pixel sizes on AP; (b) Effect of Flip Padding pixel sizes on AP; (c) Effect of Black Padding pixel sizes on R@1; (d) Effect of Flip Padding pixel sizes on R@1.}
\label{fig:shifting rubost}
\end{figure}

\textbf{(6) Robustness of MFAF to position shifting.} In this study, two position shifting, namely Black-Padding (BP) and Flip-Padding (FP), are adopted to evaluate the robustness of the proposed MFAF method. As shown in Figure \ref{fig:padding}, BP introduces a black padding region of width P to the left side of the image, cropping an equal-width strip from the right. In contrast, FP applies a mirrored extension of width P to the left edge, with a corresponding crop on the right. To validate the robustness of MFAF, comparative analyses were conducted against two method, MCCG and FSRA. The experimental results in Table \ref{tab:pad_shift} and Figure \ref{fig:shifting rubost} reveal that MFAF exhibits a markedly slower decline in accuracy relative to the competing methods with increasing padding width, indicating superior resilience to spatial shifts.

\section{Conclusion}
This paper addresses the challenge of UAV cross-view geo-localization by proposing the Multi-scale Frequency Attention Fusion (MFAF) method, which tackles the issues arising from viewpoint-induced appearance discrepancies and insufficient feature discernibility. Leveraging the EVA02 model as the backbone, MFAF demonstrates superior capability in capturing global contextual information compared to conventional CNNs and ViTs. To enable efficient integration of multi-scale low- and high-frequency features, the proposed Multi-Frequency Branch-wise Block (MFB) preserves stable global structural information through low-frequency representations, while high-frequency components capture discriminative local details. The synergistic interaction between these frequency domains substantially reduces feature inconsistencies caused by viewpoint variations. Moreover, the Frequency-aware Spatial Attention (FSA) module dynamically assigns adaptive weights to salient regions within frequency information, thereby enhancing feature discrimination while suppressing interference from irrelevant background elements. Experimental evaluations on three public datasets validate the generalizability and effectiveness of MFAF. Additionally, MFAF exhibits robust resilience to position shifting and variations in input resolution, indicating substantial practical applicability. Future research will focus on extending MFAF to more complex and challenging scenarios, including urban canyons and disaster-stricken areas, to further enhance its adaptability and resilience in diverse real-world environments.

\bibliographystyle{unsrtnat}
\bibliography{main}  






\end{document}